\documentclass[11pt]{article}

\usepackage[margin=1in]{geometry}

\newcounter{notecounter}
\usepackage{framed}
\usepackage[utf8]{inputenc} 
\usepackage[T1]{fontenc}    
\usepackage{hyperref}       
\usepackage{nccmath}
\usepackage{url}
\usepackage{soul}
\usepackage{wrapfig}
\usepackage{booktabs}       
\usepackage{amsfonts}       
\usepackage{nicefrac}       
\usepackage{microtype}      
\usepackage{xcolor}         
\usepackage{multicol}
\usepackage{graphicx}
\usepackage{amsthm}
\usepackage{bbm}
\usepackage{comment}
\usepackage{enumitem}
\usepackage{bm}
\usepackage{booktabs}
\usepackage{mathtools}
\usepackage[export]{adjustbox}
\usepackage{stmaryrd}
\usepackage{booktabs} 
\usepackage{xifthen}
\usepackage{mathtools}
\usepackage{amsmath,amssymb}
\usepackage{algorithm,algorithmic}
\usepackage{hyperref}
\usepackage{caption}
\usepackage{cleveref}
\usepackage{titlesec}
\usepackage{multirow}
\titleformat*{\subparagraph}{\itshape}

\numberwithin{equation}{section}

\usepackage{amsmath}
\usepackage{amssymb}
\usepackage{mathtools}
\usepackage{amsthm}
\usepackage{upgreek}

\newenvironment{fmtext}{}{}

\makeatletter
\let\@address\@empty
\let\@subject\@empty
\let\@keywords\@empty
\let\@corres\@empty
\newcommand{\address}[1]{\def\@address{#1}}
\newcommand{\subject}[1]{\def\@subject{#1}}
\newcommand{\keywords}[1]{\def\@keywords{#1}}
\newcommand{\corres}[1]{\def\@corres{#1}}
\renewcommand{\maketitle}{%
  \begin{center}
    {\LARGE\bfseries \@title \par}
    \vskip 1em
    {\large \@author \par}
    \vskip 0.75em
    \ifx\@address\@empty\else{\small \@address \par}\fi
    \ifx\@subject\@empty\else{\small \textbf{Subjects:} \@subject \par}\fi
    \ifx\@keywords\@empty\else{\small \textbf{Keywords:} \@keywords \par}\fi
    \ifx\@corres\@empty\else{\small \@corres \par}\fi
  \end{center}
  \vskip 1.5em
}
\makeatother
\def\rset{\mathbb{R}}

\def\rmd{\mathrm{d}}

\def\normpdf{\mathrm{N}}
\def\eqsp{\,}
\def\Id{\mathrm{I}}
\def\zero{0}

\def\pE{\mathbb{E}}

\def\var{v}
\def\eqdef{\vcentcolon=}

\def\wrt{w.r.t.}

\def\normconst{\operatorname{Z}}

\newcommand{\argmin}{\mbox{argmin}}

\def\txts{\textstyle}

\newcounter{hypA}

\newcommand{\intset}[2]{\llbracket #1, #2 \rrbracket}
\newcommand{\kldivergence}[2]{\mathsf{KL}(#1 \parallel #2)}

\def\topx{\overline{x}}
\def\botx{\underline{x}}

\def\topX{\overline{X}}
\def\botX{\underline{X}}

\def\stdobs{\sigma_\obs}
\def\obs{y}
\def\gauss{\mathcal{N}}
\def\target{\pi^\obs}
\def\param{\theta}
\def\prior{p_{\tiny{\mbox{data}}}}
\def\bprior{\bar{p}_{\tiny{\mbox{data}}}}
\def\dimobs{{d_\obs}}
\def\dimx{d}
\def\dimbot{{\underline{d}}}

\def\fwmodel{\mathcal{A}}
\newcommand{\acp}[1]{\alpha_{#1}}
\newcommand{\tbwmean}[1]{\overline{x}^\param _{#1}}

\newcommand{\score}[1]{s _{#1}}
\newcommand{\pscore}[1]{s^\param _{#1}}
\newcommand{\fw}[3]{\ifthenelse{\equal{#3}{}}{q _{#1}}{q _{#1}(#3|#2)}}
\newcommand{\fwmarg}[2]{\pdata{#1}{}{#2}}
\newcommand{\fwtrans}[3]{\ifthenelse{\equal{#2}{}}{q_{#1}}{q _{#1}(#3|#2)}}
\newcommand{\tfw}[3]{\ifthenelse{\equal{#2}{}}{\overline{q}_{#1}}{\overline{q} _{#1}(#3|#2)}}
\newcommand{\bfw}[3]{\ifthenelse{\equal{#2}{}}{\underline{q}_{#1}}{\underline{q} _{#1}(#3|#2)}}

\newcommand{\epart}[2]{\xi_{#1}^{#2}}
\newcommand{\ewght}[2]{\omega_{#1}^{#2}}

\newcommand{\ind}[2]{\kappa_{#1}^{#2}}
\def\wasser{\operatorname{W}} 
 
\newcommand{\cov}{\mbox{Cov}}

\def\ropt{r^{\tiny{\mbox{opt}}}}
\def\tropt{\overline{r}^{\tiny{\mbox{opt}}}}
\def\adjopt{\varphi^{\tiny{\mbox{opt}}}} 
\newcommand{\wgtfunc}[1]{\tilde{w}_{#1}}

\newcommand\pdata[3]{
    \ifthenelse{\equal{#2}{}}{
        \ifthenelse{\equal{#3}{}}{
            p _{#1}
        }{
            p _{#1}(#3)
        }
    }{
        \ifthenelse{\equal{#3}{}}{
            p _{#1}(\cdot|#2)
        }{
            p _{#1}(#3|#2)
        }
    }
}

\newcommand\ppdata[3]{
    \ifthenelse{\equal{#2}{}}{
        \ifthenelse{\equal{#3}{}}{
            p^\param _{#1}
        }{
            p^\param _{#1}(#3)
        }
    }{
        \ifthenelse{\equal{#3}{}}{
            p^\param _{#1}(\cdot|#2)
        }{
            p^\param _{#1}(#3|#2)
        }
    }
}

\newcommand\denoiser[3]{
    \ifthenelse{\equal{#2}{}}{
        \ifthenelse{\equal{#3}{}}{
            \operatorname{D}_{#1}
        }{
            \operatorname{D} _{#1}(#3)
        }
    }{
        \ifthenelse{\equal{#3}{}}{
            \operatorname{D} _{#1}(\cdot|#2)
        }{
            \operatorname{D} _{#1}(#3|#2)
        }
    }}

\newcommand\pdenoiser[3]{
    \ifthenelse{\equal{#2}{}}{
        \ifthenelse{\equal{#3}{}}{
            \operatorname{D}^{\param}_{#1}
        }{
            \operatorname{D}^{\param} _{#1}(#3)
        }
    }{
        \ifthenelse{\equal{#3}{}}{
            \operatorname{D}^{\param} _{#1}(\cdot|#2)
        }{
            \operatorname{D}^{\param} _{#1}(#3|#2)
        }
    }}

\newcommand\post[3]{
    \ifthenelse{\equal{#2}{}}{
        \ifthenelse{\equal{#3}{}}{
            \pi^{\obs} _{#1}
        }{
            \pi^{\obs} _{#1}(#3)
        }
    }{
        \ifthenelse{\equal{#3}{}}{
            \pi^{\obs} _{#1}(\cdot|#2)
        }{
            \pi^{\obs} _{#1}(#3|#2)
        }
    }
}

\newcommand\hpost[3]{
    \ifthenelse{\equal{#2}{}}{
        \ifthenelse{\equal{#3}{}}{
            \hat\pi^{\obs} _{#1}
        }{
            \hat\pi^{\obs} _{#1}(#3)
        }
    }{
        \ifthenelse{\equal{#3}{}}{
            \hat\pi^{\obs} _{#1}(\cdot|#2)
        }{
            \hat\pi^{\obs} _{#1}(#3|#2)
        }
    }
}

\newcommand\ihpost[4]{
    \ifthenelse{\equal{#2}{}}{
        \ifthenelse{\equal{#3}{}}{
            \hat\pi^{#4} _{#1}
        }{
            \hat\pi^{#4} _{#1}(#3)
        }
    }{
        \ifthenelse{\equal{#3}{}}{
            \hat\pi^{#4} _{#1}(\cdot|#2)
        }{
            \hat\pi^{#4} _{#1}(#3|#2)
        }
    }
}

\newcommand{\pot}[2]{
    \ifthenelse{\equal{#2}{}}{
        g _{#1}(\obs | \cdot)
        }{
            g _{#1}(\obs | #2)}
        }

\newcommand{\hpot}[2]{
    \ifthenelse{\equal{#2}{}}{
        \hat{g} _{#1}(\obs | \cdot)
        }{
            \hat{g} _{#1}(\obs|#2)}
        }

\newcommand{\ihpot}[3]{
    \ifthenelse{\equal{#2}{}}{
        \hat{g}^{#3} _{#1}(\obs | \cdot)
        }{
            \hat{g}^{#3} _{#1}(\obs|#2)}
        }

\newcommand{\hguid}[2]{
    \ifthenelse{\equal{#2}{}}{
        \hat\psi^{\obs} _{#1}
        }{
        \hat\psi^{\obs} _{#1}(#2)}
        }

\newcommand{\guid}[2]{
    \ifthenelse{\equal{#2}{}}{
        \psi^{\obs} _{#1}
        }{
        \psi^{\obs} _{#1}(#2)}
        }

\newcommand\revised[1]{#1}

\newcommand\bpdata[3]{
    \ifthenelse{\equal{#2}{}}{
        \ifthenelse{\equal{#3}{}}{
            \underline{p} _{#1}
        }{
            \underline{p} _{#1}(#3)
        }
    }{
        \ifthenelse{\equal{#3}{}}{
            \underline{p} _{#1}(\cdot|#2)
        }{
            \underline{p} _{#1}(#3|#2)
        }
    }
}

\newcommand\tpdata[3]{
    \ifthenelse{\equal{#2}{}}{
        \ifthenelse{\equal{#3}{}}{
            \overline{p} _{#1}
        }{
            \overline{p} _{#1}(#3)
        }
    }{
        \ifthenelse{\equal{#3}{}}{
            \overline{p} _{#1}(\cdot|#2)
        }{
            \overline{p} _{#1}(#3|#2)
        }
    }
}

\begin{document}

\title{Briding Diffusion Posterior Sampling and Monte Carlo methods: a survey}

\author{
Y. Janati$^{1}$, A. Durmus$^{2}$, J.~Olsson$^{3}$ and E. Moulines$^{4}$}

\address{$^{1}$Institute of Foundation Models Paris, MBZUAI\\
$^{2}$KTH Royal Institute of Technology,\\
$^{3}$Ecole polytechnique,\\
$^{4}$MBZUAI}


\newcommand\blfootnote[1]{%
  \begingroup
  \renewcommand\thefootnote{}\footnote{#1}%
  \addtocounter{footnote}{-1}%
  \endgroup
}
\blfootnote{\texttt{$\{$yazid.janati, eric.moulines$\}$@mbzuai.ac.ae, jimmyol@kth.se, alain.durmus@polytechnique.edu}}

\maketitle

\begin{abstract}
	Diffusion models enable the synthesis of highly accurate samples from complex distributions and have become foundational in generative modeling. Recently, they have demonstrated significant potential for solving Bayesian inverse problems by serving as priors. This review offers a comprehensive overview of current methods that leverage \emph{pre-trained} diffusion models alongside Monte Carlo methods to address Bayesian inverse problems without requiring additional training. We show that these methods primarily employ a \emph{twisting} mechanism for the intermediate distributions within the diffusion process, guiding the simulations toward the posterior distribution. We describe how various Monte Carlo methods are then used to aid in sampling from these twisted distributions. 
\end{abstract}


\begin{fmtext}
\section{Posterior sampling for denoising diffusion priors}
\label{sec:post_sampling_denoising}
\subsection{Bayesian inverse problems}
Inverse problems arise when we try to infer unknown causes from observed effects, which are typically determined by a forward model. 
Inverse problems are ubiquitous and play a crucial role in numerous applications across various fields, including weather forecasting, oceanography, speech processing, and image processing, especially   medical imaging (tomography) and astrophysical image reconstruction  (\emph{e.g.} black-hole imaging \cite{wu2024principled}). They share a common mathematical framework across different domains: the goal of an inverse problem is to reconstruct an 
unknown 
\end{fmtext}

\hspace{-5mm} $\mathbb{R}^{\dimx}$-valued signal $X$ from a $\mathbb{R}^{\dimobs}$-valued measurement $Y$ based on a corruption model given by 
\begin{equation} \label{eq:corruption:model}
Y = \fwmodel(X) + \sigma Z, \quad Z \sim \mathcal{N}\left(0, \Id_{\dimobs}\right) \eqsp,
\end{equation}
where \(\fwmodel\) is the forward operator, $Z$ is unobserved noise, and \(\sigma > 0\) denotes the noise level. Well-known examples include: (i) \emph{denoising}, where $\fwmodel(x)=x$ and $\sigma > 0$ and the noise does not need to be Gaussian but can be heavy-tailed or even multiplicative, as in Poisson imaging; (ii) \emph{linear inverse problems}, where $\fwmodel$ is a linear operator, \emph{i.e.} $\fwmodel(x) = A x$ for some non-invertible \(\dimobs \times \dimx\) matrix $A$ with $\dimobs \leq \dimx$, which includes, as special cases,  \emph{inpainting}, where \(A\) is a masking matrix with \(A_{ij} = 0\) for \(i \neq j\) and \(A_{ii}\) is either 0 or 1, depending on whether the corresponding entry is observed, and \emph{deblurring}, where $A$ is a 2D convolution operator (point spread function); (iii) \emph{phase retrieval}, where the goal is to recover the complex phase of a linear invertible transform from its complex modulus, \emph{i.e.} $\fwmodel(x) = |A x|$, where $A$ is an invertible matrix. The phase-retrieval framework encompasses various settings, including
\emph{image restoration} (such as super-resolution, JPEG dequantisation, and tomography) \cite{marinescu2021bayesian,soh2022variational,sahlstrom2023utilizing}, motif scaffolding in protein design \cite{trippe2023diffusion,wu2023practical}, bandwidth extension in audio-processing (reconstruction of the high-frequency components of an audio signal from its low-frequency components) \cite{Lemercier2024DiffusionMF}, and conditional trajectory simulation for urban mobility \cite{jiang2023motiondiffuser}, among others.

A key feature of these problems is that they are \emph{ill-posed}, which means that some information is lost, making exact recovery impossible. In the Bayesian inverse problem framework, the solution is treated as a \emph{distribution}. This approach represents 
unknown parameters as a random variables and updates the beliefs based on the observed data using Bayes' theorem. The Bayesian framework enables the incorporation of prior knowledge and the quantification of uncertainty, which is essential for addressing ill-posed problems. In fact, there is generally not a single solution that is compatible with the observations, but rather a set of solutions whose plausibility is measured by the posterior distribution; see \cite{stuart2010inverse,Calvetti2014Inverse,Beskos2016Geometric,calvetti2018inverse}. 

In the Bayesian approach to inverse problems, we provide the unknown signal $X$ with a prior distribution $p$.  Given the model \eqref{eq:corruption:model} and a realization $Y = y$, the likelihood function is expressed as 
$p(y|x) \propto \exp( - \| y - \fwmodel(x) \|^2 / {2\sigma^2})$. The goal is to sample from the law of $X$ given $Y = y$, which is the posterior distribution and is given by $p(x | y) \propto p(y | x)p(x)$. 
This survey focuses on a specific class of methods that employ denoising diffusion models (DDMs) as priors. These models serve as powerful priors as they can be trained to encode rich structural information about the data distribution, which can be particularly useful in settings of high-dimensional and complex data. The goal is to sample reconstructions from the posterior $p(x | y)$ using models pre-trained with data from a given distribution, without the need for retraining or fine-tuning \cite{song2019generative,ho2020denoising,song2021score,song2021ddim}.

\revised{DDMs offer a far more expressive prior than Gaussian or sparsity-based priors. Traditional Gaussian and wavelet-sparsity priors impose simplistic parametric forms or piecewise-constant structures, which limit their ability to capture the complex nature of natural images 
\cite{wang2024integrating}. In contrast, DDMs learn rich, high-dimensional data densities, enabling them to represent intricate textures, structures, and globally consistent features of images. Empirically, DDM priors lead to  reconstructions with higher fidelity. For example, \cite{feng2023score} reports that using a score-based diffusion prior yields posterior samples with significantly better SSIM/PSNR quality than those obtained with total variation or PCA-based Gaussian priors. DDMs have an exceptional ability to generate high-fidelity, realistic samples drawn from the learned distribution \cite{ho2020denoising}, whereas Gaussian priors often produce overly smooth estimates and sparsity priors can introduce piecewise-artifact or oversimplified textures.}

\revised{A major advantage of Bayesian inference with DDM priors is the ability to sample diverse solutions consistent with the measurements. Traditional priors typically lead to a single maximum a posteriori (MAP) estimate, and even when sampling is possible (\emph{e.g.}, when the posterior is Gaussian), the produced samples  mostly resemble mild noise around the MAP. In contrast, DDM priors encourage a full posterior exploration. They can generate multiple plausible reconstructions that all fit the observed data, revealing different modes of uncertainty. 
This diversity arises from the ability of DDMs to represent complex, multimodal probability densities.
As a concrete example, a diffusion prior applied to image denoising yields a family of sharp, realistic outputs, all consistent with the noisy input, whereas a sparsity prior would typically yield a single denoised image. In fact, \emph{Denoising Diffusion Restoration Models} (DDRM) \cite{kawar2022denoising}
 demonstrate that leveraging a diffusion prior enables superior reconstruction quality and diversity on challenging datasets (ImageNet), outperforming classical unsupervised methods in both fidelity and perceptual quality, while also being computationally efficient.}
 
The setup considered in this survey enables the use of a given pre-trained DDM as a prior to solve arbitrary Bayesian inverse problems, which  eliminates the need to train task-specific conditional models from scratch. This is a departure from conventional conditional DDM frameworks \cite{song2021score,batzolis2021conditional,tashiro2021csdi, saharia2022image}, which rely on paired datasets $(X, Y)$ to learn parametric approximations of the posterior distribution of the signal $X$ given the measurement $Y$.

\begin{figure}[t]
  \centering
  \includegraphics[width=.7\textwidth]{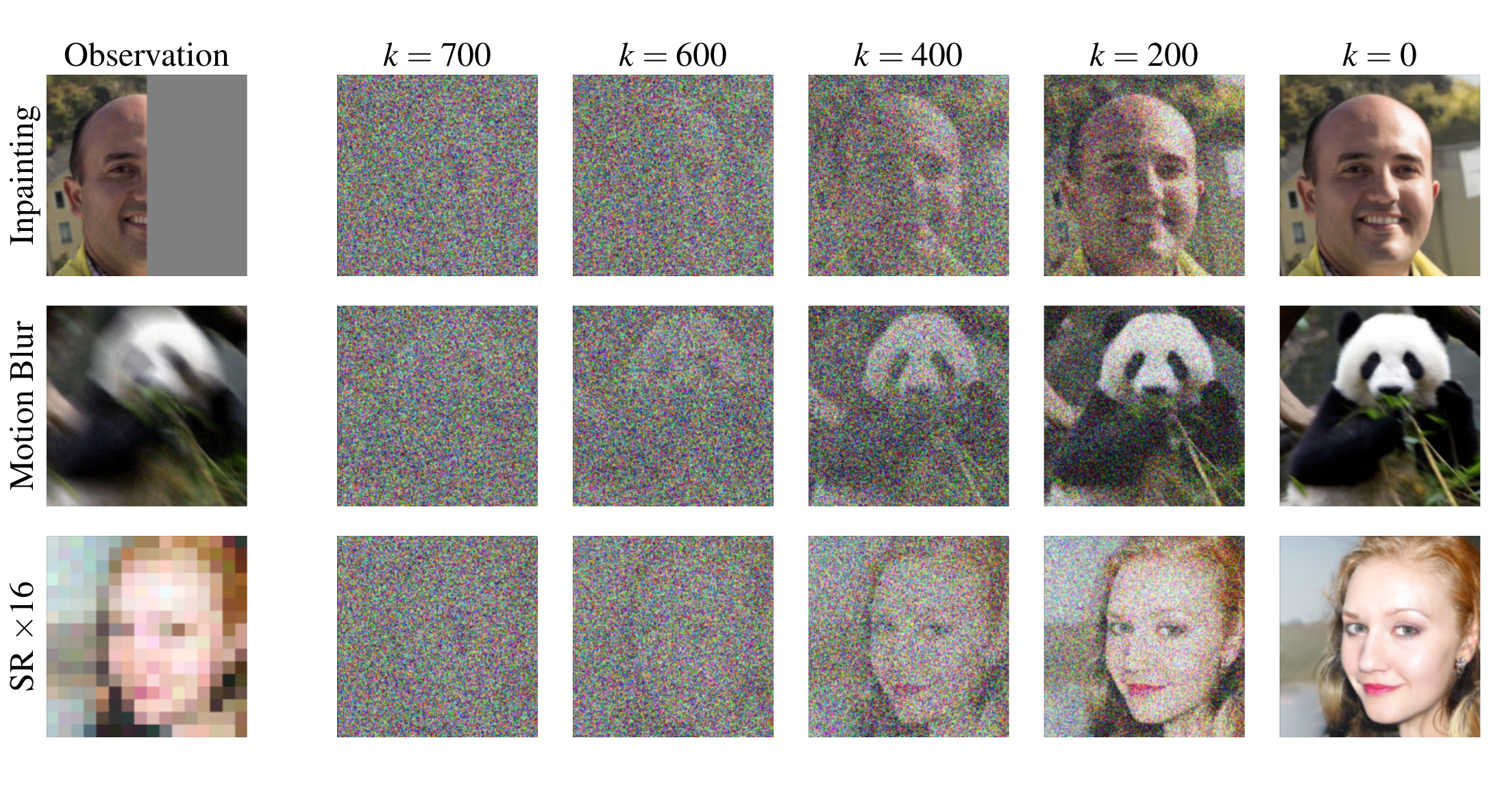}
  \caption{Progressive reconstruction of degraded images using a diffusion model. The figure demonstrates the denoising process over time ($k$ steps) for three different tasks: inpainting (top row), motion deblurring (middle row), and super-resolution (bottom row). As the diffusion process unfolds, the model progressively refines the image, recovering details from an initial noisy state to a high-quality reconstruction.}
  \label{fig:illustration}
\end{figure}

More precisely, we focus on the general problem of approximately sampling from a given target distribution
\begin{equation}  
  \label{eq:posterior_def}
   \txts \post{}{}{\rmd x} \eqdef \pot{}{x} \, \prior(\rmd x) / \normconst 
  \end{equation}
  on $\rset^{\dimx}$, where $\pot{}{}$ is some non-negative function on $\rset^{\dimx}$ referred to as the \emph{potential}, which is assumed to be evaluable pointwise,  
$\prior$ is a prior distribution, and $\normconst \eqdef \int \pot{}{x} \, \prior(\rmd x)$ is the normalising constant. When $\post{}{}{}$ is the posterior of a signal corrupted according to the model \eqref{eq:corruption:model} with positive noise level, the potential $\pot{}{}$ plays the role of the likelihood function $p(\obs | \cdot)$, where $\obs$ is the noisy observation. In contrast, in the noiseless case considered in \Cref{sec:replacement-method}, $\pot{}{}$ is rather a delta function (see \Cref{sec:replacement-method} for precise definitions). \revised{Figure~\ref{fig:illustration} depicts the typical reconstruction process of the diffused-based posterior sampling methods to be discussed next. The illustration demonstrates how these methods iteratively refine an initial noisy estimate, progressively restoring the underlying image structure.}

In parallel with this review, two complementary overviews on the same topic appeared, Daras et al. \cite{darassurvey} and Zhao et al. (2025) \cite{zhao2025conditional}; we recommend these as complementary reading.

\subsection{Denoising Diffusion models}
\label{sec:ddm}
The core idea of generative modelling is to transform an easily sampled reference distribution—such as a multivariate normal distribution—into an approximation of a data distribution, \(\prior\), which is not explicitly known, but from which we have access to samples. In DDMs \cite{sohl2015deep,song2019generative,ho2020denoising,song2020score}, the generative model is a parametric approximation \(\ppdata{0}{}{}\) of \(\prior\) on \(\rset^{\dimx}\) that is defined as the time-zero marginal of a backward Markov chain  with transition kernels \((\ppdata{k|k+1}{}{})_{k=n-1}^{0}\) and initial distribution $\ppdata{n}{}{}$; that is, \(\ppdata{0}{}{}(\rmd x_0) = \int \ppdata{0:n}{}{}(\rmd (x_0,x_{1:n}))\), where 
\begin{equation} \label{eq:ddpm-joint}
\ppdata{0:n}{}{}(\rmd x_{0:n}) \eqdef \ppdata{n}{}{}(\rmd x_n) \prod_{k = 0}^{n-1} \ppdata{k|k+1}{x_{k+1}}{\rmd x_k}
\end{equation}
denotes the joint distribution of the chain.
In this work, we set \(\ppdata{n}{}{} \eqdef \normpdf(\cdot; \zero, \Id_d)\), where $\normpdf(\cdot; \mu, \Sigma)$ is the p.d.f. of a Gaussian distribution with mean $\mu$ and covariance $\Sigma$ and 
for each $k$, 
$\ppdata{k|k+1}{x_{k+1}}{\cdot}$ 
is chosen to be a Gaussian distribution with p.d.f. 
\begin{equation}
  \label{eq:ddpm-kernel}
\ppdata{k|k+1}{x_{k+1}}{x_k} \eqdef \normpdf(x_k; \pdenoiser{k|k+1}{}{x_{k+1}}, \var^\param _{k|k+1} \Id_\dimx) \eqsp, \quad \var^\param _{k|k+1} \in \rset_{>0} \eqsp.
\end{equation}
Here, the function \(\pdenoiser{k|k+1}{}{}\) is typically a deep neural network with parameter $\theta$, which takes as input both the index \(k+1\)—usually encoded via positional encoding—and the previous state \(x_{k+1}\) (see \cite{nichol2021improved} and the references therein).   
In the context of denoising diffusion probabilistic models (DDPM) \cite{ho2020denoising}, the joint model \(\ppdata{0:n}{}{}\) is then trained to approximate the forward joint distribution
\begin{equation}
\label{eq:ddpm-forward}
\pdata{0:n}{}{\rmd x_{0:n}} = \prior(\rmd x_0) \prod_{k=0}^{n-1} \fwtrans{k+1|k}{x_k}{\rmd x_{k+1}}\eqsp,
\end{equation}
where the forward---`noising'---Markov kernels are defined as
\[
\fwtrans{k+1|k}{x_k}{x_{k+1}} \eqdef \normpdf(x_{k+1}; (\acp{k+1}/\acp{k})^{1/2} x_k, \var_{k+1|k} \Id_d)\eqsp,
\]
with \(\var_{k+1|k} \eqdef 1 - \acp{k+1}/\acp{k}\) and \((\acp{k})_{k=0}^n\) being a non-increasing sequence, with \(\acp{0} = 1\) and \(\acp{n}\) close to zero. The distribution $\fwmarg{0:n}{}$ corresponds to the joint distribution of a forward autoregressive Markov chain $(X_k)_{k = 0}^n$ with Gaussian increments and initial distribution given by the data distribution $\prior$ of interest, \emph{i.e.},
\begin{equation}
\label{eq:noising_process}
    X_{k+1} = (\alpha_{k+1}/\alpha_k)^{1/2} X_k + v_{k+1|k} Z_{k+1} \eqsp, \qquad X_{0}
 \sim \prior  \eqsp, \end{equation}
 where $(Z_k)_{k=1}^n$ are 
 independent and standard Gaussian.
 Although many alternative noising and backward kernels have been proposed, we follow the approach in \cite{ho2020denoising} for ease of presentation.

Set \(\var_{k|0} \eqdef 1 - \acp{k}\) and denote by
\begin{equation}
\label{eq:forward-marginal}
\pdata{k}{}{\rmd x_k} \eqdef \int \fwtrans{k|0}{x_0}{\rmd x_k} \, \prior(\rmd x_0) , \quad \text{where} \quad \fwtrans{k|0}{x_0}{x_k} \eqdef \normpdf(x_k; \sqrt{\acp{k}} x_0, \var_{k|0} \Id_\dimx)\eqsp, 
\end{equation}
the 
\(k\)-th marginal of \eqref{eq:ddpm-forward}. Note that since $\acp{n}$ is supposed to be almost zero, $\pdata{n}{}{}$ is approximately a standard Gaussian distribution, which justifies the definition of $\ppdata{n}{}{}$. Defining the backward Markov kernels 
\begin{equation}
\label{eq:backward-markov-kernel}
\pdata{k|k+1}{x_{k+1}}{\rmd x_k} \eqdef \fwtrans{k+1|k}{x_k}{x_{k+1}} \pdata{k}{}{\rmd x_k}  / \pdata{k+1}{}{x_{k+1}} 
\end{equation}
allows the joint distribution \eqref{eq:ddpm-forward} to be alternatively expressed as
\begin{equation}
  \label{eq:ddpm-backward}
  \pdata{0:n}{}{\rmd x_{0:n}} = \pdata{n}{}{\rmd x_n} \prod_{k = 0}^{n-1} \pdata{k|k+1}{x_{k+1}}{\rmd x_k} \eqsp.
\end{equation}
This \emph{backward decomposition} is primarily of theoretical interest, as the backward Markov kernels \eqref{eq:backward-markov-kernel} are not tractable due to their dependence on the marginal distributions $\pdata{k}{}{}$, which are unknown.  Still, it justifies the form of the model  \eqref{eq:ddpm-joint}--\eqref{eq:ddpm-kernel}. Indeed, 
note first that for every $x_{k+1}$, 
the mean $\pE_{\pdata{0:n}{}{}}[X_k | X_{k+1} = x_{k+1}]$ and the covariance $\cov_{\pdata{0:n}{}{}}(X_k | X_{k+1} = x_{k+1})$ minimise the Kullback--Leibler divergence
$
(\mu,\Sigma) 
\mapsto 
\kldivergence{ \pdata{k|k+1}{x_{k+1}}{\cdot}}{\normpdf(\cdot ; \mu,\Sigma)}.
$
  Here, the subscript $\prior$ indicates that the mean and the covariance both are taken with respect to the joint distribution $\pdata{0:n}{}{}$ (see \eqref{eq:ddpm-backward}).
Therefore, for each DDPM backward transition density \eqref{eq:ddpm-kernel}, the mapping $\pdenoiser{k|k+1}{}{}$  parameterising the mean is a neural network approximating 
$x_{k+1} \mapsto \pE_{\pdata{0:n}{}{}}[X_k | X_{k+1} = x_{k+1}]$, while the covariance $\var_{k|k+1}^{\theta} \Id_\dimx$ is chosen such that $\var_{k|k+1}^{\theta}= v_{k|0,k+1} := \var_{k|0}\var_{k+1|k}/v_{k+1|0}$, which is---as shown below---a lower bound on the covariance of the exact time-reversed process.  

Indeed, it is easily shown that under the joint distribution $\fwmarg{0:n}{}$, the conditional density of $X_k$ given $X_{k+1} = x_{k+1}$ and $X_0 = x_0$ is 
\begin{equation}
  \label{eq:bridge-kernel}
  \fwtrans{k|0, k+1}{x_0, x_{k+1}}{x_k} \eqdef \normpdf(x_k; \gamma_k \acp{k}^{1/2} x_0 + (1 - \gamma_k) (\acp{k+1} / \acp{k})^{-1/2} x_{k+1}, \var_{k|0, k+1} \Id_d) \eqsp,
\end{equation}
with $\gamma_k \eqdef \var_{k+1|k} / \var_{k+1|0}$.
It then follows that 
\begin{align}
  \label{eq:expectation-decomp}
  \pE_{\pdata{0:n}{}{}}[X_k | X_{k+1}] & = \gamma_k \acp{k}^{1/2} \pE_{\pdata{0:n}{}{}}[X_0 | X_{k+1}] + (1 - \gamma_k) (\acp{k+1} / \acp{k})^{-1/2} X_{k+1} \eqsp, \\
  \label{eq:covariance-decomp}
  \mathrm{Cov}_{\pdata{0:n}{}{}}(X_k | X_{k+1}) & = \gamma^2 _k \acp{k} \mathrm{Cov}_{\pdata{0:n}{}{}}(X_0 | X_{k+1}) + \var_{k| 0, k+1} \Id_d \eqsp. 
\end{align}
Drawing inspiration from these identities, \cite{ho2020denoising} considers the parameterisation
$$
\pdenoiser{k|k+1}{}{x_{k+1}} =   \gamma_k \acp{k}^{1/2} \pdenoiser{k+1}{}{x_{k+1}} + (1 - \gamma_k) (\acp{k+1} / \acp{k})^{-1/2} x_{k+1} \eqsp,
$$ 
where $\smash{\pdenoiser{k+1}{}{}}$ is 
a denoiser approximating the mapping $x_{k + 1} \mapsto \pE_{\pdata{0:n}{}{}}[X_0 | X_{k+1} = x_{k + 1}]$,
and sets $\smash{\var^\param _{k|k+1}} = \var_{k|0, k+1}$. 
As a result, each DDPM transition \eqref{eq:ddpm-kernel} can be written in the more compact form
\begin{equation}
  \label{eq:ddpm-backward}
  \ppdata{k|k+1}{x_{k+1}}{x_k} = \fwtrans{k|0, k+1}{}{}(x_k | \pdenoiser{k+1}{}{x_{k+1}}, x_{k+1}) \eqsp.
\end{equation}

An important feature of DDMs is that learning the denoisers also yields parametric approximations of the Stein scores $\nabla \log \fwmarg{k}{}$, which explains the term `score-based generative modelling' used in the literature \cite{song2021score}. Indeed,  Tweedie's formula \cite{robbins1956empirical} yields
\begin{equation}
  \label{eq:tweedie}
  \nabla \log \fwmarg{k}{x_k} = \big(- x_k + \sqrt{\acp{k}}\, \pE_{\pdata{0:n}{}{}}[X_0 | X_k = x_k]\big) / \var_{k|0} \eqsp,  
\end{equation}
and therefore we obtain a parametric approximation $\pscore{k}$ of $\nabla \log \fwmarg{k}{}$ by replacing, in the previous identity, $\pE_{\pdata{0:n}{}{}}[X_0 | X_k = \cdot]$ by the approximate denoiser $\pdenoiser{k}{}{}$. 

In the remainder of the paper, we denote by \((\ppdata{k}{}{})_{k=0}^n\) the marginals of the joint model \eqref{eq:ddpm-joint}, defined 
by $\ppdata{n}{}{} \eqdef \normpdf(\cdot; \zero, \Id_d)$ and, recursively, 
\begin{equation}
\label{eq:definition-bwp-k}
\ppdata{k}{}{}(x_k) \eqdef \int \ppdata{k|k+1}{x_{k+1}}{x_k} \, \ppdata{k+1}{}{}(\rmd x_{k+1}) \eqsp.
\end{equation}
For all $0 \leq \ell < m \leq n$ we also define the multi-step transition densities 
\begin{equation}
  \label{eq:ddpm-conditional}
\ppdata{\ell|m}{x_m}{\rmd x_\ell} \eqdef \int \prod_{k = \ell}^{m-1} \ppdata{k|k+1}{x_{k+1}}{\rmd x_k} \eqsp.
\end{equation}
Similarly, we let $\pdata{\ell|m}{x_m}{\rmd x_\ell}$ be the multi-step transition defined for the transition kernels \eqref{eq:backward-markov-kernel}. We also denote by $\denoiser{k}{}{}$ the true denoiser mapping $x_k \mapsto \pE _{\pdata{0:n}{}{}}[X_0 | X_k = x_k]$ and by $\score{k}$ the true score $\nabla \log \pdata{k}{}{}$. 

\subsection{Posterior sampling}
\revised{
A DDM prior inherently imposes  constraints that limit flexibility when constructing samplers for the target posterior distribution. Indeed, recall that the posterior distribution $\post{}{}{}$, defined in \eqref{eq:posterior_def}, explicitly depends on the prior $\prior$, which is typically intractable. In practice, we only have access to a parametric approximation, denoted $\ppdata{0}{}{}$, which corresponds to the marginal distribution at time zero of the backward diffusion process described by the joint model \eqref{eq:ddpm-joint}. As a result, computing the exact density of $\ppdata{0}{}{}$ through numerical integration of the associated probability-flow ordinary differential equation (ODE) \cite{song2021score} is computationally expensive. Moreover, accurately estimating its score function $\nabla \log \ppdata{0}{}{}$, even when it is well-defined, is generally infeasible. These computational challenges make standard sampling techniques, such as \emph{Markov chain Monte Carlo} (MCMC) or \emph{sequential Monte Carlo} (SMC), impractical for posterior inference under a DDM prior.}
Still, we have access to pre-trained score approximations \(\pscore{k}\), denoisers \(\pdenoiser{k}{}{}\), and the marginals \(\ppdata{k}{}{}\) of \eqref{eq:ddpm-joint}. The challenge, therefore, lies in combining these components to construct an efficient sampler for the posterior distribution.

From a theoretical perspective, the DDM approach for sampling from the target distribution \(\post{}{}{}\) is equivalent to simulating a backward Markov chain \((X_k^{\obs})_{k=n}^0\), where each state \(X_k^{\obs}\) is (approximately) distributed according to 
\(\post{k}{}{}\), defined by
\begin{equation}
\label{eq:target-posterior}
  \post{k}{}{\rmd x_k} \eqdef \int \fwtrans{k|0}{x_0}{\rmd x_k} \, \post{}{}{\rmd x_0} = \int \fw{k|0}{x_0}{\rmd x_k} \frac{\pot{}{x_0}  \prior(\rmd x_0)}{\prior(\pot{}{})} 
  \end{equation}
  for $k \geq 1$ and 
  with $\post{0}{}{}$ being identical with the target posterior \(\post{}{}{}\). Here we used the short-hand notation $\prior(\pot{}{})= \int  \pot{}{x_0} \, \prior(\rmd x_0)$. Note that \eqref{eq:target-posterior} is analogous to \eqref{eq:forward-marginal}, but with the crucial difference that \(\prior\) is replaced by \(\target\). The backward transition kernels are given by 
\begin{equation}
    \label{eq:posterior-backward}
\post{k|k+1}{x_{k+1}}{\rmd x_k} \eqdef \frac{\post{k}{}{\rmd x_k} \fwtrans{k+1|k}{x_k}{x_{k+1}}}{\post{k+1}{}{x_{k+1}}} \eqsp,
\end{equation}
where \(\fwtrans{k+1|k}{}{}\) is the forward kernel. 
From \eqref{eq:posterior-backward} we immediately obtain the recursion $\post{k}{}{\rmd x_k} = \int \post{k|k+1}{x_{k+1}}{\rmd x_k} \, \post{k+1}{}{\rmd x_{k+1}}$. \revised{This suggests that, given an approximate sample \( X_{k+1}^{\obs} \) from the posterior distribution \(\post{k+1}{}{}\), drawing a subsequent sample \( X_{k}^{\obs} \) from a parametric approximation of the backward kernel \(\post{k|k+1}{X_{k+1}^{\obs}}{\cdot}\) yields an approximate sample from \(\post{k}{}{}\). As detailed in the preceding section, the DDM approximation of \(\post{k|k+1}{}{}\) requires parametric approximations of the posterior score functions \(\left(\nabla \log \post{k}{}{}\right)_{k=1}^n\), which are derived via Tweedie's formula; see \eqref{eq:tweedie}. Specifically, these posterior scores can be expressed in terms of the prior scores \(\left(\nabla \log \pdata{k}{}{}\right)_{k=1}^n\), augmented by an additional \emph{guidance term}.} Indeed, we can relate \(\post{k}{}{}\) to \(\fwmarg{k}{}\) through the expression 
\begin{equation}
\label{eq:posterior-forward}
\post{k}{}{\rmd x_k} = \frac{\pot{k}{x_k} \fwmarg{k}{\rmd x_k}}{\fwmarg{k}{g_k(\obs | \cdot)}}\eqsp,
\end{equation}
where 
\[
\begin{split}
\pot{k}{x_k} &\eqdef \int \pot{}{x_0} \, \pdata{0|k}{x_k}{\rmd x_0} \eqsp, \\ 
\pdata{0|k}{x_k}{\rmd x_0} &\eqdef \prior(\rmd x_0) \, \fw{k|0}{x_0}{\rmd x_k} \big/ \fwmarg{k}{x_k}\eqsp.
\end{split}
\]

The structure of \(\post{k}{}{}\) mirrors that of the posterior distribution \(\target\), in that the former is the product of the forward process's \(k\)-th marginal, \(\fwmarg{k}{}\), and a potential function that can be interpreted as the likelihood of the observation given the forward process at time \(k\).
The score is then given by 
\begin{equation}
\label{eq:conditional-score}
\nabla \log \post{k}{}{x_k} = \nabla \log \fwmarg{k}{x_k} + \nabla \log \pot{k}{x_k} \eqsp.
\end{equation}
The guidance term \(\nabla \log \pot{k}{x_k}\) quantifies how the original score should be modified to skew the sampling procedure towards the posterior distribution. Although a parametric approximation is readily available for the first term appearing on the right-hand side of \eqref{eq:conditional-score}, 
the principal difficulty arises from the second term, which involves integrating the potential function \(\pot{}{}\) with respect to the conditional distribution \(\pdata{0|k}{x_k}{}\). This integral is generally intractable, and accurately approximating this quantity is challenging, constituting the primary computational bottleneck in the method.

In the following sections, we discuss how existing approaches have  addressed the inherent intractability of the potential function \eqref{eq:posterior-forward} and the associated conditional scores \eqref{eq:conditional-score}, resulting in efficient, \emph{learning-free} algorithms for sampling from the posterior distribution \(\pi\). 


\section{Gradient guidance}
\label{sec:dps-type-approx}

In what follows, we discuss several methods designed to sample from the posterior distributions \((\post{k}{}{})_{k=n}^0\) defined in \eqref{eq:posterior-denoiser}, by recursively approximating the backward transition densities given in \eqref{eq:posterior-backward}. Extending the approach outlined in \Cref{sec:post_sampling_denoising}\ref{sec:ddm}, but now 
letting 
\(\target\) play the role of \(\prior\)---and 
consequently replacing \(\fwmarg{k}{}\) by \(\post{k}{}{}\) for each \(k \geq 1\)---requires approximating the conditional denoising function 
\((x_k, k) \mapsto \pE_{\post{0:n}{}{}}[X_0 \mid X_k = x_k]\). This conditional expectation is computed under the conditional distribution 
\(\post{0|k}{x_k}{\rmd x_0} \propto \post{}{}{\rmd x_0}\,\fwtrans{k|0}{x_0}{x_k}\). Finally, combining expressions \eqref{eq:tweedie} and \eqref{eq:conditional-score}, we obtain that
\begin{equation}
    \label{eq:posterior-denoiser}
    \pE_{\post{0:n}{}{}} [X_0 | X_k = x_k] = \pE_{\pdata{0:n}{}{}}[X_0 | X_k = x_k] + \frac{v_{k|0}}{\sqrt{\acp{k}}} \nabla_{} \log \pot{k}{x_k} \eqsp,  
\end{equation}
where $\pot{k}{}$ is defined in \eqref{eq:posterior-forward}. Thus, if we then have access to (unconditional) pre-trained denoisers $(\denoiser{k}{}{}{})_{k = 1} ^n$, we can use various methods, which will be discussed next, to approximate each conditional denoiser 
and enable approximate sampling from $\post{}{}{}$ using a DDPM-like approximation. 
From now on, we use the short-hand notation $\guid{k}{} \eqdef \nabla \log g_k(\obs|\cdot)$ to denote the guidance term at step $k$. 
By combining the definition of the DDPM approximation (given jointly by \eqref{eq:bridge-kernel}--\eqref{eq:ddpm-backward}) with \eqref{eq:posterior-denoiser}, we obtain that, given $X_{k + 1}$, an approximate draw $X_k$ from \(\post{k|k+1}{X_{k+1}}{\cdot}\) can be obtained by, first, drawing $\widetilde{X}_k$ from $ \pdata{k|k+1}{X_{k+1}}{\cdot}$ and, second, letting  
\begin{equation}
\label{eq:ddpm-posterior-update}
    X_k = \widetilde{X}_k + 
    \var_{k+1|k}(1 - \var_{k+1|k})^{-1/2}
    \guid{k+1}{X_{k + 1}}
    \eqsp.
\end{equation}
As already mentioned, however, both $\pot{k}{}$ and $\guid{k}{}$ are generally intractable; therefore, we must  find surrogates \(\hpot{k}{}\) or $\hguid{k}{}$ for one of these quantities. \revised{We denote by $\pot{k}{}$ and $\guid{k}{}$ these parametric surrogates.} 
The methods that we will discuss below draw inspiration from \eqref{eq:ddpm-posterior-update} and apply updates in the form 
\begin{equation}
\label{eq:dps-update}
    X_k = \tilde{X}_{k} + \wgtfunc{k}(X_{k+1}) \hguid{k+1}{X_{k+1}}\eqsp,
\end{equation}
where \(\wgtfunc{k}\) is a weight function. The methods differ in the design of the score approximation $\hguid{k+1}{}$ and the weight function $\wgtfunc{k}$, and we next provide a description of the most common approaches.  

The \emph{Diffusion Posterior Sampling} (DPS) algorithm proposed in \cite{chung2023diffusion} is one of the most popular posterior sampling algorithms for DDM-prior-based Bayesian inverse problems. This method makes use of the approximation 
\begin{equation}
    \label{eq:dps-approximation}
  \hguid{k}{x_{k}} = 
  \nabla_{x_k} \log g(\obs | \denoiser{k}{}{x_k})
  = \nabla_{x_k} \denoiser{k}{}{x_k} ^\intercal \nabla_{x_0} \log \pot{}{x_0}_{|x_0 = \denoiser{k}{}{x_k}} \eqsp,
    \end{equation} 
which amounts to replacing $\pdata{0|k}{x_k}{\rmd x_0}$ by $\delta_{\denoiser{k}{}{x_k}}(\rmd x_0)$ in the definition \eqref{eq:posterior-forward}.  Notably, it implies in practice the computation of a vector-Jacobian product involving the denoiser neural network. When $\pot{}{x} = \normpdf(\obs; \fwmodel(x), \stdobs^2 \Id_{\dimobs})$ (see \eqref{eq:corruption:model}), the update \eqref{eq:dps-update} is used with 
$$
\wgtfunc{k}(x_{k+1}) = \gamma \stdobs^2 / \| \obs - \fwmodel(\denoiser{k+1}{}{x_{k+1}}) \| \eqsp, 
$$
where $\gamma \in [0,1]$, which can be seen as a way to stabilise the updates by taming the gradient.

The \emph{Pseudo-Inverse Guided Diffusion Model} ($\Pi$GDM) developed in \cite{song2022pseudoinverse}, which generalises \cite{ho2022video}, approximates each transition density $\pdata{0|k}{x_k}{\cdot}$ by $\normpdf(\cdot; \denoiser{k}{}{x_k}, \var_{0|k} \Id)$, where the variance \(\var_{0|k}=1 - \acp{k}\) matches that of \(\pdata{0|k}{x_k}{x_0}\) under the assumption that \(\prior\) is standard Gaussian. In the specific case of linear inverse problems, where $\pot{}{x} = \normpdf(\obs; Ax, \Sigma_\obs)$ and $\Sigma_\obs$ is a positive definite matrix, this approximation leads to the guidance-term approximation
\begin{equation}
    \label{eq:pgdm-approximation}
    \hguid{k}{x_{k}} = \nabla \log \normpdf(\obs; A \denoiser{k}{}{x_k}, \var _{0|k} AA^\intercal + \Sigma_\obs) \eqsp.
\end{equation}
We emphasise that the covariance is not related to the actual data distribution. In this case, the weight function used in the update \eqref{eq:dps-update} is $w_k(\cdot) \equiv \sqrt{\acp{k+1}} \var _{0|k+1}$. 


The \emph{moment matching algorithm} introduced in \cite{finzi2023user,boys2023tweedie} aims to 
solve the optimisation problem
\begin{equation}
\label{eq:gaussian-approximation}
\argmin_{(\mu,\Sigma) \in \rset^\dimx \times \mathcal{S}_{++}^{\dimx}} \kldivergence{\normpdf(\cdot;\mu,\Sigma)}{\pdata{0|k}{x_k}{\cdot}}\eqsp,
\end{equation}
where \(\mathcal{S}_{++}^{\dimx}\) denotes the set of \(\dimx \times \dimx\) symmetric positive definite matrices. The optimal solution \((\mu_{0|k}, \Sigma_{0|k})\) is explicitly given by
\begin{align}
    \mu_{0|k}(x_k) & = \pE_{\pdata{0:n}{}{}} [X_0 \mid X_k = x_k]\eqsp,\quad \nonumber \\ 
    \Sigma_{0|k}(x_k) & = \operatorname{Cov}_{\pdata{0:n}{}{}} (X_0 \mid X_k = x_k)  
    = \frac{\var_{k|0}}{\sqrt{\acp{k}}}\nabla \pE_{\pdata{0:n}{}{}} [X_0 \mid X_k = x_k]\eqsp,
    \label{eq:cov-jac}
\end{align}
where the last equality is derived in \cite{meng2021estimating,boys2023tweedie}.
Hence, given a pre-trained denoiser $\denoiser{k}{}{}$, we can obtain the Gaussian approximation $\normpdf (x_0; \denoiser{k}{}{x_k}, (\var_{k|0} / \sqrt{\acp{k}}) \nabla \denoiser{k}{}{x_k})$ of $\pdata{0|k}{x_k}{x_0}$. For linear inverse problems, the exact integration of the likelihood $\pot{}{}$ w.r.t. the Gaussian approximation \eqref{eq:gaussian-approximation}, treating the resulting covariance matrix as a \emph{constant function} of $x_k$ and using that $\nabla \denoiser{k}{}{}$ is symmetric following \eqref{eq:cov-jac}, yields the guidance approximation
\begin{equation}
    \label{eq:tmpd-grad}
   \hguid{k}{x_k} = \nabla \denoiser{k}{}{x_k} A^\intercal \bigg[ \frac{\var_{k|0}}{\sqrt{\acp{k}}} A \nabla \denoiser{k}{}{x_k} A^\intercal + \Sigma_\obs \bigg]^{-1} \big(\obs - A \denoiser{k}{}{x_k}\big) \eqsp.
\end{equation}
When plugging in the parametric approximation $\pdenoiser{k}{}{}$, this approach requires computing the Jacobian matrix of a neural network, which  can become prohibitively expensive. 
Moreover, it also entails a computationally costly matrix inversion, further adding to the complexity.
It is important to emphasise that since \( \denoiser{k}{}{x_k} \) is estimated, its Jacobian \(\nabla \denoiser{k}{}{x_k}\) is not necessarily positive semi-definite. To mitigate these computational issues, \cite{boys2023tweedie} proposes a computationally lighter diagonal approximation, obtained by replacing the Jacobian matrix in \eqref{eq:dps-update} with its diagonal approximation defined by the vector-Jacobian product \(\nabla \denoiser{k}{}{x_k}^\intercal \mathbf{1}_\dimx\). This vector can be computed efficiently using standard automatic-differentiation libraries, thereby substantially reducing computational overhead. 

Alternatively, \cite{rozet2024learning} seeks to circumvent the expensive evaluation of the guidance term \(\hguid{k}{}\) defined in \eqref{eq:tmpd-grad}. The proposed method employs the conjugate gradient algorithm to efficiently solve the linear system
\[
M_k z = y - A\denoiser{k}{}{x_k}\eqsp,\quad z \in \rset^\dimobs\eqsp,
\]
where the matrix \(M_k\) is defined as
\[
M_k \eqdef \var_{k|0}\acp{k}^{-1/2}A\nabla \denoiser{k}{}{x_k} A^\intercal + \Sigma_\obs\eqsp.
\]
This strategy avoids explicit computation and inversion of high-dimensional matrices, providing a scalable alternative suitable for practical applications. If we denote the solution by $z^*$, we obtain \eqref{eq:tmpd-grad} by differentiating $\denoiser{k}{}{x_k}^\intercal A^\intercal z^*$ while treating $z^*$ as a constant. This computation is relatively inexpensive since it reduces to a vector-Jacobian product. To solve the system using conjugate gradient, it is only necessary to calculate products $M_k v$ for $v \in \rset^\dimobs$, which can be done efficiently using vector-Jacobian products. This method allows \eqref{eq:tmpd-grad} to be computed without any Jacobian calculation or explicit matrix inversion.  \cite{boys2023tweedie,rozet2024learning} both use 
 $\wgtfunc{k}(\cdot) \equiv \var_{k+1|k}(1 - \var_{k+1|k})^{-1/2}$ 
 as in the update \eqref{eq:ddpm-posterior-update}.
  
\section{A sequence-of-distributions perspective}
\label{sec:sequence-distribution}

Guidance-based sampling methods inherently introduce systematic bias due to the approximation of the true potential functions \(\pot{k}{}\) by surrogate functions \(\hpot{k}{}\). To mitigate this bias, recent research proposes substituting the original sequence of posterior distributions \((\post{k}{}{})_{k=n}^0\) with an alternative sequence \((\hpost{k}{}{})_{k=n}^0\). This alternative sequence is carefully designed to achieve a smooth interpolation from the initial Gaussian distribution \(\hpost{n}{}{} = \gauss(0, \Id_d)\) to the target posterior distribution \(\hpost{0}{}{} = \post{}{}{}\), while remaining suitable for efficient simulation using standard Monte Carlo techniques \cite{robert2004monte}.

An appealing strategy for constructing such a sequence of intermediate distributions consists in replicating the structure of \(\post{k}{}{}\) as defined in \eqref{eq:posterior-forward}, but replacing the true potential \(\pot{k}{}\) with a surrogate potential \(\hpot{k}{}\). Notably, the surrogate potential \(\hpot{k}{}\) does not need to be a precise approximation to \(\pot{k}{}\). More specifically, we define the modified distributions by
\begin{equation}
    \label{eq:generic-distr}
    \hpost{k}{}{\rmd x_k} \propto \hpot{k}{x_k} \pdata{k}{}{\rmd x_k}\eqsp.
\end{equation}
We emphasise, however, that alternative constructions for these intermediate sequences are possible, as discussed further in \Cref{sec:sequence-distribution}\ref{subsubsec:mcmc} and \Cref{sec:inpainting}\ref{subsec:replacement-method}.
Before exploring how standard sampling methods, such as \emph{sequential Monte Carlo} (SMC) and \emph{Markov chain Monte Carlo} (MCMC), have been applied in the literature to distributions of the form \eqref{eq:generic-distr}, we first discuss some common choices for the surrogate potentials \(\hpot{k}{}\). A prominent example, studied in  \cite{zhang2023towards, wu2023practical} and more recently in \cite{zhu2024think}, is the DPS approximation  given by 
\begin{equation}
\label{eq:dps-approximation}
\hpot{k}{x_k} = g(\obs | \denoiser{k}{}{x_k}) . 
\end{equation}
In contrast, the simpler approach of setting  \(\hpot{k}{} = \pot{}{}\) has been proposed in \cite{wu2024principled,xu2024provably}.  For Gaussian likelihood, \(\pot{}{x} = \normpdf(\obs; A x, \stdobs^2 \Id_\dimobs)\), the following alternative approximation, inspired by the \(\Pi\)GDM methodology and \eqref{eq:pgdm-approximation}, was suggested in \cite{rozet2023score}:
\[
\hpot{k}{x_k} = \normpdf\left(\obs; A \denoiser{k}{}{x_k}, \gamma \frac{1 - \acp{k}}{\acp{k}} A A^\intercal + \stdobs^2 \Id_\dimobs\right). 
\]
For linear inverse problems, several works have explored Gaussian potentials with means that are linear in the variable \(x_k\). Notable examples include \cite{jalal2021robust,cardoso2024monte,dou2024diffusion,janati2024divide}. In particular, \cite{janati2024divide} proposes a surrogate potential given by
\(
\hpot{k}{x_k} = \normpdf(\sqrt{\acp{k}}\,\obs; A x_k, \stdobs^2 \Id_\dimobs).
\)
\subsection{Sequential Monte Carlo methods}
\label{subsec:smc}
SMC (sequential Monte Carlo) methods aim to sample from a generic sequence \((\hpost{k}{}{})_{k=n}^0\) of distributions that are generated by a recursion of the form 
\begin{equation}
\label{eq:FK-semigroup}
\hpost{k}{}{\rmd x_k} = \frac{1}{\normconst_k}\int m_{k|k+1}(\rmd x_k \mid x_{k+1}) w_{k}(x_k,x_{k+1}) \hpost{k+1}{}{\rmd x_{k+1}},
\end{equation}
where \(w_k\) is a non-negative weight function, $\normconst_k$ 
is a normalising constant, 
and \(m_{k|k+1}\) is a Markov kernel; see \cite{gelman1998simulating, delmoral2006smcs, dai2022invitation}.
This recursion is often referred to as a non-linear Feynman--Kac semigroup and is studied in detail in, \emph{e.g.}, \cite{del2004feynman}. We insist that there are many possible choices of  $(w_k, m_{k|k+1})_{k = 0} ^{n-1}$ such that \eqref{eq:FK-semigroup} holds for a given  distribution sequence $(\hpost{k}{}{})_{k = n} ^0$.
Since 
\(\pdata{k}{}{}(x_k) = \int \pdata{k|k+1}{x_{k+1}}{x_k} \pdata{k+1}{}{\rmd x_{k+1}}\), each \(\hpost{k}{}{}\) defined in \eqref{eq:generic-distr} can be expressed as the marginal of the joint distribution
\begin{equation}
    \label{eq:posterior-joint}
    \hpost{k, k+1}{}{\rmd x_{k: k+1}} \propto \hpot{k}{x_k} \pdata{k|k+1}{x_{k+1}}{\rmd x_k} \pdata{k+1}{}{\rmd x_{k+1}} 
\end{equation}
w.r.t. $x_k$. From this it follows that the distribution flow $(\hpost{k}{}{})_{k = n}^0$ satisfies \eqref{eq:FK-semigroup} with 
\begin{equation}
    \label{eq:recursion-example}
    m_{k|k+1}(\rmd x_k| x_{k+1})= \pdata{k|k+1}{x_{k+1}}{\rmd x_k} \eqsp, \quad w_k(x_k,x_{k+1}) = \frac{\hpot{k}{x_k}}{\hpot{k+1}{x_{k+1}}} \eqsp. 
\end{equation}
\revised{In practice, when needed, the  transition kernels and weight functions \eqref{eq:recursion-example} are replaced with their parametric surrogates $\ppdata{k|k+1}{x_{k+1}}{}$ and $(x_k, x_{k+1}) \mapsto \hpot{k}{x_k} / \hpot{k+1}{x_{k+1}}$. }
There are various approaches for sampling from a sequence of distributions of the form \eqref{eq:FK-semigroup}. 
The most straightforward approach is to use SMC methods (also known as particle filters) which approximate each distribution \(\hpost{k}{}{} \) with a set $(\epart{k}{i})_{i=1}^N$ of random draws referred to as particles. The particles are corrected using non-negative importance weights $(\ewght{k}{i})_{i=1}^N$. More specifically, the self-normalised weighted empirical measure $\sum_{i=1}^N \ewght{k}{i} \delta_{\epart{k}{i}} /  \sum_{i=1}^N \ewght{k}{i}$ serves as an approximation of $\hpost{k}{}{}$. 

An SMC algorithm recursively propagates the weighted particles from one time step $k$ to the next using importance sampling and resampling techniques, providing an approximation of \(\hpost{k}{}{}\).
We refer to \cite{del2004feynman,chopin2020introduction} for comprehensive treatments of the theory and practice of SMC methods.

To illustrate how a weighted particle sample can be evolved through the Feynman--Kac recursion \eqref{eq:FK-semigroup}, suppose we currently have a particle sample \((\ewght{k+1}{i},\epart{k+1}{i})_{i=1}^N\) with associated weights, such that the empirical distribution \(\ihpost{k+1}{}{}{N}\) approximates the posterior \(\hpost{k+1}{}{}\).

In order to construct an updated particle approximation \((\ewght{k}{i},\epart{k}{i})_{i=1}^N\) of \(\hpost{k}{}{}\), we substitute the empirical measure \(\ihpost{k+1}{}{}{N}\) into the recursion \eqref{eq:FK-semigroup}. This substitution yields a weighted mixture distribution:
\[
\bar{\pi}_k^N(\rmd x_k) \propto \sum_{i = 1}^N \ewght{k + 1}{i} w_k(x_k, \epart{k + 1}{i})\, m_{k\mid k+1}(\rmd x_k \mid \epart{k + 1}{i}).
\]
In the next step, new particles are sampled from $\bar{\pi}_k^N$ using importance sampling based on some mixture proposal $\rho_k^N(\rmd x_k) \propto \sum_{i = 1}^N \varphi_{k + 1}^i \ewght{k + 1}{i} \, r_{k|k+1}(\rmd x_k \mid \epart{k + 1}{i})$, where $r_{k|k+1}$ is some instrumental transition density dominating $m_{k|k+1}$ and $(\varphi_{k + 1}^i)_{i = 1}^N$ is a set of positive weights referred to as adjustment multipliers \cite{pitt1999filtering}. 
In fact, to keep the computational cost down, one samples rather from  an extension $\Pi _k^N(i, \rmd x_k) \propto \ewght{k + 1}{i} w_k(x_k, \epart{k + 1}{i}) \, m_{k|k+1}(\rmd x_k \mid \epart{k + 1}{i})$ of $\bar{\pi}_k^N$ to the product space $\{1, \ldots, N\} \times \rset^{\dimx}$ of mixture indices and particle locations using a similar extension of $\rho_k^N$. 
More precisely, this importance sampling procedure boils down to: (i) drawing indices $(\ind{k}{\ell})_{\ell = 1}^N$ (conditionally) independently from the categorical distribution on $\{1, \ldots, N\}$ with probability vector proportional to $(\ewght{k + 1}{\ell} \varphi_{k + 1}^\ell)_{\ell = 1}^N$; (ii) generating independently, for all sampled indices, new particles $\epart{k}{i} \sim r_{k|k+1}(\rmd x_k \mid \epart{k + 1}{\ind{k}{i}})$; (iii) assigning each new particle $\epart{k}{i}$ an updated weight $\ewght{k}{i} \eqdef w_k(\epart{k}{i}, \epart{k + 1}{\ind{k}{i}}) (\rmd m_{k|k+1} /\rmd r_{k|k+1})(\epart{k}{i}, \epart{k + 1}{\ind{k}{i}}) / \varphi^{\ind{k}{i}}_k$, where $\rmd m_{k|k+1} / \rmd r_{k|k+1}$ denotes the Radon--Nikodym derivative of $m_{k|k+1}$ w.r.t. $r_{k|k+1}$. 
In the end, we have generated updated particle samples $(\epart{k}{i}, \ewght{k}{i})_{i = 1}^N$ approximating $\hpost{k}{}{}$, and by repeating recursively this process, starting with a uniformly weighted sample of particles drawn from the standard Gaussian distribution, a particle approximation $(\epart{0}{i}, \ewght{0}{i})_{i = 1}^N$ of the posterior $\pi$ of interest is obtained. 

The sampling of the mixture indices in (i) can be equivalently understood as resampling, with replacement, new particles among the old ones in proportion to the adjusted weights, whereas the sampling step (ii) serves to jitter, or shake, the particles randomly. We will therefore refer to operations (i--iii) collectively as the \textsc{resample--shake--weight} scheme (in the literature, operations (i) and (ii) are also often referred to as selection and mutation, respectively).

The performance of the \textsc{resample--shake--weight} scheme depends mainly on the choice of the proposal density and the adjustment multipliers. If the particles are mutated `blindly', without regard to the shape of $w_k$, the resulting particle weights will likely be significantly skewed, leading to high variance. 

In some situations, it is possible to sample from the  \emph{optimal} proposal kernel $\ropt_{k|k+1}(\rmd x_k \mid x_{k + 1}) \propto w_k(x_k, x_{k + 1}) m_{k|k+1}(\rmd x_k \mid x_{k + 1})$ and also calculate the optimal adjustment multipliers $\adjopt_{k + 1}(x_{k + 1}) = \int w_k(x_k, x_{k + 1}) \, m_{k|k+1}(\rmd x_k \mid x_{k + 1})$. In that case, we can sample exactly from $\bar{\pi}_k^N$, without the need of importance sampling, by instead: (i') computing the weights $(\adjopt_{k+1}(\epart{k + 1}{i}) \ewght{k + 1}{i})_{i = 1}^N$; (ii') drawing independently mixture indices $(\ind{k}{\ell})_{\ell = 1}^N$ from the categorical distribution on $\{1, \ldots, N\}$ induced by the weights in (i'); (iii') generating new particles $\epart{k}{i} \sim \ropt_{k|k+1}(\rmd x_k \mid \epart{k + 1}{\ind{k + 1}{i}})$. Appealingly, this procedure, which we refer to as the \textsc{weight--resample--shake} scheme, will always generate uniformly weighted particles at all iterations. 

The \emph{Twisted Diffusion Sampler} (TDS) proposed in
\cite{wu2023practical} targets a sequence of distributions in the form \eqref{eq:generic-distr} with potentials $\hpot{k}{x_k} = g(\obs|\denoiser{k}{}{x_k})$ by constructing empirical approximations of each distribution in accordance with the previous scheme. As instrumental kernel \(r_{k|k+1}\) the authors suggest using the one corresponding to the update \eqref{eq:dps-update}, where \(\guid{k+1}{}\) is approximated by \eqref{eq:dps-approximation}, and the weight function $\wgtfunc{k}$ is constant and equal to  
\(\var_{k+1|k}(1 - \var_{k+1|k})^{-1/2}\). 
In this case, it holds that $w_k(x_k,x_{k+1})= g(\obs|\denoiser{k}{}{x_k}) / g(\obs|\denoiser{k+1}{}{x_{k+1}})$ and $\rmd m_k/\rmd r_k (x_k,x_{k+1}) = \ppdata{k|k+1}{x_{k+1}}{x_k}/ r_k(x_k \mid x_{k+1})$ and the authors use the \textsc{resample--shake--weight} scheme. 
Note, however, that the gradient of the denoiser $\nabla \denoiser{k}{}{}$ must be calculated at each iteration and for each particle, which can make TDS computationally intensive.

The \emph{Monte Carlo Guided Diffusion} (MCGDiff) algorithm,  introduced by \cite{cardoso2024monte}, represents another SMC approach specifically tailored to linear inverse problems. MCGDiff employs a sequence of distributions 
\eqref{eq:generic-distr} defined by the Gaussian potentials 
\begin{equation}
    \label{eq:mcgdiff-pot}
\hpot{k}{x_k} = \normpdf(\sqrt{\acp{k}}\obs; A x_k, \Sigma_k)\eqsp,
\end{equation}
where \(\Sigma_k\) is a positive definite covariance matrix. In this case, due to the linear dependence of \(\hpot{k}{}\) on \(x_k\), both the optimal proposal kernel 
\begin{equation} 
    \label{eq:optimal-transition}
\ropt_{k|k+1}(x_k \mid x_{k+1}) \propto \hpot{k}{x_k}\,\pdata{k|k+1}{x_{k+1}}{x_k}
\end{equation}
and the corresponding optimal adjustment multiplier function
\[
\adjopt_{k+1}(x_{k+1}) = \frac{\int \hpot{k}{x_k}\,\pdata{k|k+1}{x_{k+1}}{\rmd x_k}}{\hpot{k+1}{x_{k+1}}}
\]
\revised{can be computed exactly upon replacing $\pdata{k|k+1}{x_{k+1}}{}$ with its  parametric approximation $\ppdata{k|k+1}{x_{k+1}}{}$.} 
Exploiting this tractability, MCGDiff employs a \textsc{weight–resample–shake} SMC scheme, resulting in significantly lower computational costs compared to the TDS method. Further details on the choice and construction of the potentials in MCGDiff can be found in \Cref{sec:inpainting}\ref{subsec:mcgdiff}. 

A related approach for linear inverse problems was proposed in \cite{dou2024diffusion}, which introduced  random Gaussian potentials given by 
\[
\hpot{k}{x_k} = \normpdf(Y_k; A x_k, \acp{k}\stdobs^2 \Id_\dimobs)\eqsp,
\]
where \(Y_k\) is a \emph{pseudo-observation} drawn from the marginal distribution
\(\gauss(\sqrt{\acp{k}}\obs, (1 - \acp{k})\Id_\dimobs)\). In this approach, the authors use the optimal proposal kernel
\[
r_{k|k+1}(x_k \mid x_{k+1}) \propto \hpot{k}{x_k}\,\pdata{k|k+1}{x_{k+1}}{x_k}\eqsp,
\]
while adopting adjustment multiplier functions \(\varphi_{k+1}(x_{k+1}) = \int \hpot{k}{x_k}\,\pdata{k|k+1}{x_{k+1}}{\rmd x_k}\) that differ from the optimal choice. Nevertheless, after the resampling and shaking steps, the resulting particle weights are reset to unity.

\revised{SMC methods represent an attractive and theoretically well-founded class of algorithms for posterior sampling, owing particularly to their capability to generate particle approximations endowed with strong theoretical guarantees (see, \emph{e.g.}, \cite{douc2008limit,delmoral:2013}). Practical applications of SMC methods can, however, face significant challenges. 
A major limitation arises in high-dimensional settings, where \emph{particle-weight degeneracy} becomes a prevalent issue. This leads to a highly skewed weight distribution among the particles, causing increased variance and reduced effectiveness of the approximation.
To alleviate degeneracy and maintain satisfactory approximation accuracy, it becomes necessary to increase the particle population size \(N\), often dramatically \cite{bickel:li:bengtsson:2008}. However, scaling up the number of particles drastically inflates the computational cost, potentially resulting in prohibitive memory usage 
and extended runtimes. 
}

\revised{Moreover, strategies such as adaptive resampling and particle rejuvenation, while helpful, may only partially mitigate this issue. Consequently, in very high-dimensional or computationally intensive inference tasks---common in fields like imaging, machine learning, or inverse problems---alternative approaches, such as tailored MCMC methods, hybrid methods, or dimension-reduction strategies, may be preferred in practice.}

\subsection{Markov chain Monte Carlo and variational inference methods}
\label{subsubsec:mcmc}
We now circumvent the computational complexity of SMC methods by directly generating a sequence  $(\hat{X}_k)_{k=n}^{0}$ where each sample $\hat{X}_k$ is approximately distributed according to $\hpost{k}{}{}$ as defined in \eqref{eq:posterior-joint}. This approach starts by initialising $\hat{X}_n \sim \gauss(0, \Id_d)$ and then proceeds recursively using MCMC (Markov chain Monte Carlo) methods.

More specifically, for each step $k \in \intset{0}{n-1}$, we run a short homogeneous Markov chain $(\hat{X}_k^j)_{j=0}^R$ targeting the distribution $\hpost{k}{}{}$. This chain is initialised at $\hat{X}_k^0 = \hat{X}_{k+1}$, and after $R$ iterations, we set the next sample as $\hat{X}_k = \hat{X}_k^R$. While theoretically advantageous, incorporating an acceptance-rejection step based on exact density evaluations of $\hpost{k}{}{}$ is often impractical due to the intractability of the distributions given in \eqref{eq:generic-distr}.

A primary technical challenge is designing an efficient and accurate MCMC kernel that targets $\hpost{k}{}{}$, since its density, given by
\begin{equation}
\label{eq:approximate-target-hpost-k}
\hpost{k}{}{\rmd x_k} \propto \hpot{k}{x_k}\,\pdata{k}{}{\rmd x_k} \eqsp,
\end{equation}
\revised{is typically intractable, even after replacing $\pdata{k}{}{}$ with $\ppdata{k}{}{}$. }
Therefore, we can only rely on the score $\score{k}$ and its approximation. The \emph{Score Annealed Langevin Dynamics} (Score ALD) algorithm \cite{jalal2021robust}, when adapted to the DDPM framework, uses the potential
\[
\hpot{k}{x_k}= \normpdf\big(\sqrt{\acp{k}} \, \obs; A x_k, \acp{k}(\sigma_y^2 + \vartheta_k^2)\big),
\]
where $\vartheta_k^2$ is a tunable hyperparameter. Then approximate samples from $\hpost{k}{}{}$ are generated using a single iteration of the unadjusted Langevin dynamics given by
\begin{equation*}
    \hat{X}_{k}^1 = \hat{X}_{k}^0 + \eta_k \{ \score{k}(\hat{X}^0_k ) + \nabla \log \hpot{k}{\hat{X}^0_k} \} + \sqrt{2\eta_k} Z_k^1 \eqsp, 
\end{equation*}
where $\hat{X}_{k} ^0 = \hat{X}^1 _{k+1}$, $\eta_k>0$ is a step size, and $Z_k^1$ is a $d$-dimensional standard Gaussian random variable. The \emph{Score Data Assimilation} (SDA) algorithm of \cite{rozet2023score} considers instead the potential
\[
\hpot{k}{x_k} = \normpdf\big(\obs; A \denoiser{k}{}{x_k}, \Sigma_\obs + \gamma (1 - \acp{k}) A A^\intercal / \acp{k}\big),
\]
where $\gamma > 0$ is a hyperparameter, assuming $\pot{}{x} = \normpdf(\obs; A x, \Sigma_\obs)$ with $\Sigma_\obs \in \mathcal{S}_+^\dimx$. It employs a nested Markov chain $(X_k)_{k=n}^{0}$, where each intermediate chain $(X_k^i)_{i=0}^{M}$ is initialised by applying the update \eqref{eq:dps-update} with the guidance term $\hguid{k}{} = \nabla \log \hpot{k}{}$. Subsequent iterations follow Langevin-like updates with a state-dependent step size according to 
\[
X_k^{i+1} = X_k^i + \eta_k(X_k^i) \{\pscore{k}(X_k^i) + \nabla_{x_k} \log \hpot{k}{X_k^i}\} + \sqrt{2 \eta_k(X_k^i)} Z_i,
\]
where $Z_i \sim \gauss(\zero_\dimx, \Id_\dimx)$ and $\eta_k(x) \eqdef \eta \dimx / \| \pscore{k}(x) + \nabla_{x_k} \log \hpot{k}{x} \|^2_2$, with $\eta$ being a hyperparameter, gives the step size as a function of the state. Note that due to the state-dependent step sizes, these updates do not inherit the theoretical guarantees of standard Langevin dynamics. In \cite{zhu2024think}, the authors adopt a similar approach but initialise the chain unconditionally via $X_k^0 \sim \pdata{k|k+1}{X_{k+1}}{\cdot}$ and use state-independent step sizes.

Further, leveraging the fact that the posterior $\hpost{k}{}{}$ in \eqref{eq:generic-distr} is a marginal of the joint distribution \eqref{eq:posterior-joint}, we can adopt a two-stage Gibbs-sampling approach. This involves alternating between sampling the conditional distributions
\begin{align*}
    \hpost{k+1|k}{x_k}{x_{k+1}} & \eqdef \fw{k+1|k}{x_{k}}{x_{k+1}}, \\
    \hpost{k|k+1}{x_{k+1}}{x_k} & \propto \hpot{k}{x_k}\, \pdata{k|k+1}{x_{k+1}}{x_k} \eqsp.
\end{align*}
The second conditional distribution, which corresponds to the optimal kernel \eqref{eq:optimal-transition} in the SMC approach of the previous section, is generally intractable---even after substituting the parametric approximation $\ppdata{k|k+1}{x_{k+1}}{}$---unless, for instance, the potential $\hpot{k}{}$ takes the form \eqref{eq:mcgdiff-pot}.
 More generally, sampling from the second conditional $\hpost{k|k+1}{x_{k+1}}{}$ can be efficiently carried out using a Metropolis--Hastings step;  given a current state $X_k^j$, a candidate move $\tilde{X}_k^{j+1}$ is proposed using an auxiliary kernel $r^y _k(\cdot \mid X_k^j; x_{k+1})$ and accepted with  probability $\alpha_k(X^j _k, \tilde{X}^{j+1} _k)$, where 
\[
\alpha_k(x,\tilde{x}) \eqdef \min \left(1, \frac{\hpot{k}{\tilde{x}}\, \pdata{k|k+1}{x_{k+1}}{\tilde{x}}\, r^\obs _k(x|\tilde{x}; x_{k+1}) }{\hpot{k}{x}\, \pdata{k|k+1}{x_{k+1}}{x}\, r^\obs _k(\tilde{x}| x; x_{k+1})}\right).
\]
The \emph{CoPaint} algorithm introduced in \cite{zhang2023towards} is an instance of this Metropolis-within-Gibbs sampler, \revised{more specifically, when the time-travel frequency (denoted by $\tau$ in \cite[Algorithm~1 and Section~3.2]{zhang2023towards}) is set to 1, and when the auxiliary kernel $r^\obs _k(\cdot | \cdot; x_{k+1})$ is chosen as a Metropolis Adjusted Langevin Algorithm (MALA) kernel \cite{besag:1994}:
$$
    r^\obs _k(\tilde{x} | x; x_{k+1}) \eqdef \normpdf (\tilde{x}; x + \delta \{ \nabla_x \log \hpot{k}{x} + \nabla_x \log \pdata{k|k+1}{x_{k+1}}{x} \}, 2 \delta 
    \, \Id_\dimx) \eqsp, 
$$
where $\delta > 0$ is the step size. } 

The Gibbs sampler also underpins the \emph{Plug-and-Play Diffusion Model} (PnP-DM) and \emph{Diffusion Plug-and-Play} (DPnP) methods introduced in \cite{wu2024principled, xu2024provably}. However, their formulation considers a sequence of intermediate distributions $(\hpost{k}{}{})_{k=n}^0$ that differ from \eqref{eq:generic-distr}. 

\subsubsection{Split Gibbs sampling}
\label{subsubsec:split-gibbs}
Consider, for $k \in \intset{0}{n}$, sampling from $\post{k}{}{}$, defined in \eqref{eq:posterior-forward}, using Gibbs sampling. This distribution is the marginal with respect to the second variable of the joint distribution 
$$\post{0,k}{}{\rmd (x_0, x_k)} \propto \pot{}{x_0} \prior(\rmd x_0) \fwtrans{k|0}{x_0}{\rmd x_k} \propto \pot{}{x_0} \pdata{0|k}{x_k}{\rmd x_0} \fwmarg{k}{\rmd x_k}\eqsp. $$
The two conditional distributions associated with this joint distribution are explicitly given by
\[
\post{k|0}{x_0}{\rmd x_k} = \fwtrans{k|0}{x_0}{\rmd x_k}, \quad \text{and} \quad \post{0|k}{x_k}{\rmd x_0} \propto \pot{}{x_0}\,\pdata{0|k}{x_k}{\rmd x_0}.
\]
Sampling from the first conditional is straightforward, whereas sampling from the second conditional poses substantial challenges. Specifically, (i) the conditional distribution $\pdata{0|k}{x_k}{\cdot}$ typically does not possess a density with respect to the Lebesgue measure in most practical settings, and (ii) even when a density exists, accurate approximations of its scores or density are typically unavailable, thereby precluding standard sampling approaches. Consequently, $\pdata{0|k}{x_k}{\cdot}$ remains tractable exclusively for direct sampling, making approximate inference or sampling from the posterior $\post{0|k}{x_k}{\cdot}$ practically infeasible.

\revised{The methods proposed in \cite{wu2024principled, xu2024provably} are inspired by the proximal gradient method \cite{parikh2014proximal} and the split Gibbs sampling (SGS) scheme \cite{vono2019split}. These approaches construct a joint distribution specifically designed to yield fully tractable conditional distributions within Gibbs sampling. Define for $k \in \intset{0}{n}$, 
\begin{equation} 
    \label{eq:sgs}
    \hpost{k}{}{\rmd (x_0, x_k)} \propto   \pot{}{x_k} \fwtrans{k|0}{x_0}{\rmd x_k} \pdata{0}{}{\rmd x_0}
\end{equation}
with the convention that \(\fwtrans{0|0}{x_0}{\rmd x_k} = \delta_{x_0}(\rmd x_k)\).}
The full conditional distributions associated with this joint are fully tractable. Specifically, they are given by
\[
\hpost{k|0}{x_0}{\rmd x_k} \propto \pot{}{x_k}\,\fwtrans{k|0}{x_0}{\rmd x_k}, \quad \text{and} \quad \hpost{0|k}{x_k}{\rmd x_0} \eqdef \pdata{0|k}{x_k}{\rmd x_0} \eqsp. \]
 In the context of linear inverse problems with Gaussian noise, the first conditional distribution, $\hpost{k|0}{x_0}{\cdot}$, can be sampled exactly using standard Gaussian conjugacy formulas. For nonlinear inverse problems, sampling can be performed via the unadjusted Langevin algorithm as demonstrated in \cite{wu2024principled,xu2024provably}, or by employing any appropriate Markov Chain Monte Carlo (MCMC) method. As for the conditional $\post{0|k}{x_k}{}$, it can be sampled approximately by sampling from the pre-trained denoising distribution $\ppdata{0|k}{x_k}{}$ \eqref{eq:ddpm-conditional}. 

Note that the second marginal of the joint distribution \(\hpost{0, k}{}{}\) is \(\hpost{k}{}{\rmd x_k} \propto \pot{}{x_k} \pdata{k}{}{\rmd x_k}\). As a result, the split Gibbs sampling procedure described in the previous paragraph is equivalent to sampling from \(\hpost{k}{}{}\) using Gibbs sampling with the conditionals of this same joint distribution. Consequently, \cite{wu2024principled} and \cite{xu2024provably} implicitly consider distribution sequences of the form \eqref{eq:generic-distr} with \(\hpot{k}{} = \pot{}{}\).
This implies that the same sampling scheme can be applied to any sequence of the form \eqref{eq:generic-distr}, such as those discussed in the previous and upcoming sections, by simply replacing \(\pot{}{}\) with \(\hpot{k}{}\). We discuss one such related approach in the next section.

\subsubsection{Mixture Guided Posterior Sampling}
\revised{
By construction, the potential $\pot{k}{}$, defined in \eqref{eq:posterior-forward}, can be written, for any $j \in \intset{0}{k-1}$, as follows 
    $$ 
        \pot{k}{x_k} = \int \pot{j}{x_j} \pdata{j|k}{x_k}{\rmd x_j} \eqsp.
    $$ 
   This decomposition is interesting because it allows defining various approximations $\pot{k}{x_k}$ by substituting an approximation of the potential $\pot{j}{x_j}$ in the previous expression. As we will see below, a particularly fruitful idea is to use the DPS approximation \eqref{eq:dps-approximation}. Instead of selecting a particular value $j \in \intset{0}{k-1}$ it is suggested in \cite{janati2025mixture} to use all these decompositions simultaneously with non-negative weights $\{ w^j _k\}_{j=0}^{k-1}$ satisfying $\sum_{j = 1}^{k-1} w^j _k = 1$. More precisely, \cite{janati2025mixture} proposes to sample from a sequence of mixture distributions, where each mixture component is  of the form \eqref{eq:generic-distr}:
    \begin{equation*} 
        \hpost{k}{}{x_k} \eqdef  \sum_{j = 1} ^{k-1} w^j _k \ihpost{k}{}{x_k}{j} \eqsp, \quad \text{where} \quad \ihpost{k}{}{x_k}{j} \propto \ihpot{k}{x_k}{j} \pdata{k}{}{x_k}
    \end{equation*}
    and $\ihpot{k}{x_k}{j} \eqdef \int g(\obs|\denoiser{j}{}{x_j}) \pdata{j|k}{x_k}{\rmd x_j}$.}  
    
\revised{In order to approximately sample from the marginal distribution $\hpost{k}{}{}$, the \emph{Mixture Guided Diffusion Model} proposed by \cite{janati2025mixture} begins by sampling a mixture component $j \sim \mathrm{Categorical}(\{ w_k^\ell \}_{\ell = 1}^{k-1})$, followed by applying a Gibbs sampler targeting each individual component $\ihpost{k}{}{}{j}$ in the mixture. By construction of $\ihpot{k}{}{j}$, the distribution $\ihpost{k}{}{}{j}$ can be expressed as a marginal of the joint distribution
\[
    \hpost{0,j,k}{}{\rmd (x_0,x_j,x_k)} \propto \pdata{0|j}{x_j}{\rmd x_0}\,\hpot{j}{x_j}\,\pdata{j|k}{x_k}{\rmd x_j}\,\pdata{k}{}{\rmd x_k} \eqsp.
\]
The introduction of $x_0$ may seem quite artificial at this stage, but it proves particularly useful when computing the full-conditional distributions. The full conditional distributions associated with this joint distribution are explicitly given by
\begin{align*}
    \hpost{0|j,k}{x_j,x_k}{\rmd x_0} & = \pdata{0|j}{x_j}{\rmd x_0},  \\
    \hpost{j|0,k}{x_0,x_k}{\rmd x_j} & \propto \hpot{j}{x_j}\,\pdata{0|j}{x_j}{x_0}\,\pdata{j|k}{x_k}{\rmd x_j}, \\
    \hpost{k|0,j}{x_0,x_j}{\rmd x_k} & \propto \pdata{j|k}{x_j}{x_k}\,\pdata{k}{}{\rmd x_k}.
\end{align*}
Following the definition \eqref{eq:forward-marginal}, \eqref{eq:backward-markov-kernel}, the second and third full conditionals can be simplified to
\[
    \hpost{k|0,j}{x_0,x_j}{\rmd x_k} = \fw{k|j}{x_j}{\rmd x_k}. \nonumber
\]
where $\fw{k|j}{x_j}{x_k} \eqdef \normpdf(x_k; (\acp{k} / \acp{j})^{1/2} x_j, (1 - \acp{k} / \acp{j}) \Id_\dimx)$ and 
\begin{equation}
    \hpost{j|0,k}{x_0,x_k}{\rmd x_j}  = \frac{\hpot{j}{x_j}\,\fw{j|0,k}{x_0,x_k}{\rmd x_j}}{\int \hpot{j}{x_j'}\,\fw{j|0,k}{x_0,x_k}{\rmd x_j'}}, \label{eq:jth-posterior}
\end{equation}
where 
$$
    \fw{j|0, k}{x_0, x_k}{x_j} \eqdef \normpdf(x_j; \gamma_{k|j} \acp{j}^{1/2} x_0 + (1 - \gamma_{k|j}) (\acp{k} / \acp{j})^{-1/2} x_k, v_{k|0, j} \cdot \Id_\dimx) 
$$ 
and $v_{k|0, j} \eqdef (1 - \acp{j})(1 - \acp{k} / \acp{j}) / (1 - \acp{k})$. 
In this scenario, $\hpost{k|0,j}{x_0, x_j}{\cdot}$ can be drawn exactly, $\hpost{0|j,k}{x_j, x_k}{\cdot}$ can be sampled approximately using the pre-trained model but sampling from $\hpost{j|0,k}{x_0, x_k}{\cdot}$ necessitates approximate methods. To address this, \cite{janati2025mixture} propose utilizing either variational inference or employing a Metropolis--Hastings algorithm, as described in the preceding section. With the variational inference approach, the algorithm proceeds as follows. Having drawn an index $j \sim \mbox{Categorical}(\{ w^\ell _k \}_{\ell =1} ^{k-1})$ and given the previous state of the Markov chain $(X^r _0, X^r _j, X^r _k)$, the next state $(X^{r+1} _0, X^{r+1} _j, X^{r+1} _k)$ is obtained by first fitting a Gaussian variational approximation to the distribution approximation \eqref{eq:jth-posterior} of $\hpost{j|0,k}{X^r _0,X^r _k}{}$. This is done by fitting a Gaussian distribution with diagonal covariance via the minimization, using stochastic gradient descent, of the KL divergence
\begin{multline*} 
    \mathsf{KL}\left( \normpdf(\mu, \Id_\dimx v) \left\| \frac{\hpot{j}{x_j}\,\fwtrans{j|0,k}{X^r _0,X^r _k}{\rmd x_j}}{\int \hpot{j}{x_j'}\,\fwtrans{j|0,k}{X^r_0,X^r _k}{\rmd x_j'}} \right. \right) \\ 
    = - \pE_{\gauss(\zero_\dimx, \Id_\dimx)}\big[\log \hpot{j}{\mu + (\Id_\dimx v^{1/2}) Z}\big] + \kldivergence{\normpdf(\mu, \Id_\dimx v)}{\fwtrans{j|0, k}{X^r _0, X^r _k}{\cdot}} + \mathrm{C}\eqsp.
\end{multline*}
\wrt\ the mean and covariance parameters $(\mu , v) \in \rset^\dimx \times \rset^\dimx _{>0}$ and $C$ is a constant independent of these parameters. Denoting by $(\hat{\mu}_j, \hat{v}_j)$ the obtained parameters, we draw $X^{r+1} _j$ from $\gauss(\hat{\mu}_j, \Id_\dimx \hat{v}_j)$. As for the remaining states, we have that $X^{r+1} _0 \sim \pdata{0|j}{X^{r+1} _j}{}$ and $X^{r+1} _k \sim \fwtrans{k|j}{X^{r+1} _j}{\cdot}$. 
We note that the variational inference step does not rely on amortized variational approximations; instead, when generating \(N\) samples in parallel, \(N\) individual Gaussian variational approximations are fitted by optimizing both the mean and diagonal covariance vectors for each sample. A similar non-amortized approach is adopted in \cite{mardani2024a}, where Gaussian variational approximations are used to directly fit the posterior distribution \eqref{eq:posterior_def}. \\
In the next two sections, we present alternative methods that leverage the DPS approximation \eqref{eq:dps-approximation} in combination with variational inference and MCMC techniques, each in distinct ways.
} 
\subsubsection{Divide-and-Conquer Posterior Sampling (DCPS)}
\label{subsubsec:DCPS}
While the works discussed in \Cref{sec:dps-type-approx} aim to improve the approximation of the denoising distribution \(\pdata{0|k}{x_k}{\cdot}\), and thereby the approximation of \(\pot{k}{}\), \cite{janati2024divide} takes a different approach. This approach leverages the fact that approximation errors are provably smaller when targeting conditionals \(\pdata{\ell|k}{x_k}{\cdot}\) for \(\ell\) close to \(k\). Specifically, as shown in \cite[Proposition 3.1]{janati2024divide}, if \(\hat{p}_{0|k}(\cdot \mid x_k)\) approximates \(\pdata{0|k}{x_k}{}\), then the transition with density \(\hat{p}_{\ell | k}(x_\ell \mid x_k) = \int \fwtrans{\ell|0, k}{x_0, x_k}{x_\ell} \hat{p}_{0|k}(\rmd x_0 \mid x_k)\) satisfies, for all \(x_k \in \rset^\dimx\) and \(\ell \in \intset{1}{k-1}\),
\[
    \wasser_2(\pdata{\ell|k}{x_k}{\cdot}, \hat{p}_{\ell|k}(\cdot \mid x_k)) \leq C_{\ell, k} \wasser_2(\pdata{0|k}{x_k}{\cdot}, \hat{p}_{0|k}(\cdot \mid x_k))\eqsp,
\]
where \(C_{\ell, k} \eqdef \sqrt{\acp{\ell}} \left( 1 - {\acp{k}}/{\acp{\ell}} \right) / (1 - \acp{k}) < 1\), and \(\wasser_2\) denotes the $2$-Wasserstein distance. The approximate transition $\hat{p}_{\ell|k}(\cdot | x_k)$ is computable if for example $\hat{p}_{0|k}(\cdot | x_k)$ is any of the approximations discussed in \Cref{sec:dps-type-approx} such as $\hat{p}_{0|k}(\cdot | x_k) \eqdef \delta_{\denoiser{k}{}{x_k}}$. To leverage this insight the authors consider a sequence $(\hpost{k_\ell}{}{})_{\ell = L}^0$ of the type \eqref{eq:generic-distr} where $(k_\ell)_{\ell = 0} ^L$ is an increasing sequence of timesteps with $k_0 = 0,\, k_L = 1$ and $L$ is much smaller than $n$. The potentials are defined as $\hpot{k_\ell}{}: x_{k_\ell} \mapsto \pot{}{x_{k_\ell} / \eta_{k_\ell}}^{\beta_{k_\ell}}$, where $(\gamma_{k_\ell})_{\ell = L} ^0$ and $(\eta_{k_\ell})_{\ell = L}^0$ are problem specific sequences that satisfy $\eta_0 = \beta_0 = 1$. In the case of linear inverse problems they set $\eta_{k_\ell} = \sqrt{\acp{k_\ell}}$ and $\beta_{k_\ell} = \acp{k_\ell}$. Then, instead of using the recursion \eqref{eq:FK-semigroup} which calls for a particle approximation, $\hpost{k_\ell}{}{}$ at step $k_\ell$ is instead expressed as the marginal of a backward Markov chain with an initial distribution at step $k_{\ell+1}$; i.e. it holds that 
\begin{align*} 
\hpost{k_\ell}{}{\rmd x_{k_\ell}} 
& \propto \hpot{k_\ell}{x_{k_\ell}} \pdata{k_{\ell}}{}{\rmd x_{k_{\ell}}} \\
& \propto \hpot{k_\ell}{x_{k_\ell}} \int \prod_{j = k_{\ell}} ^{k_{\ell+1} - 1}  \pdata{j|j+1}{x_{j+1}}{\rmd x_j} \pdata{k_{\ell+1}}{}{\rmd x_{k_{\ell+1}}} \\
&= \int \ihpost{k_{\ell+1}}{}{\rmd x_{k_\ell + 1}}{\ell} \prod_{j = k_{\ell}} ^{k_{\ell+1} - 1} \ihpost{j|j+1}{x_{j+1}}{\rmd x_j}{\ell} 
\end{align*}
where $\ihpost{k_{\ell+1}}{}{\rmd x_{k_{\ell+1}}}{\ell} \propto \ihpot{k_{\ell+1}}{x_{k_{\ell +1 }}}{\ell} \pdata{k_{\ell+1}}{}{\rmd x_{k_{\ell+1}}}$, $\ihpost{j|j+1}{x_{j+1}}{\rmd x_j}{\ell} \propto \ihpot{j}{x_j}{\ell} \pdata{j|j+1}{x_{j+1}}{\rmd x_j}$ and for $j \in \intset{k_\ell}{k_{\ell+1}}$, 
$$
\ihpot{j}{x_j}{\ell} \eqdef \int \hpot{k_\ell}{x_{k_\ell}} \pdata{k_\ell|j}{x_j}{\rmd x_{k_\ell}} \eqsp.
$$ 
We emphasize that the initial distribution \(\smash{\ihpost{k_{\ell+1}}{}{}{\ell}}\) differs from \(\hpost{k_{\ell+1}}{}{}\), and thus, to generate an approximate sample from \(\hpost{k_\ell}{}{}\), starting from a sample of \(\hpost{k_{\ell+1}}{}{}\), one must (i) first target \(\smash{\hpost{k_{\ell+1}}{}{}{\ell}}\), (ii) followed by simulating the Markov chain with transitions \(\smash{(\ihpost{j|j+1}{}{}{\ell})_{j = k_\ell}^{k_{\ell+1} - 1}}\). However, both the initial distribution and the transitions are intractable for approximate inference, as their definition depends on the intractable potential \(\ihpot{j}{}{\ell}\). This intractability is similar to that of \(\pot{k}{}\), with the key difference being that the integral now involves \(\pdata{k_\ell|j}{x_j}{\cdot}\) instead of \(\pdata{0|j}{x_j}{\cdot}\). 
As a first step in addressing the problem of sampling from $\hpost{k_{\ell}}{}{}$, the authors introduce surrogates which consists in replacing \(\ihpot{j}{}{\ell}\) with its approximation \(\int \hpot{k_{\ell}}{x_{k_\ell}}\, \hat{p} _{k_{\ell}|j}(\rmd x_{k_\ell} \mid x_j)\), \revised{where $\pdata{k_\ell|j}{x_j}{\cdot}$ has been replaced with $\hat{p} _{k_\ell|j}(\cdot| x_j) \eqdef \fwtrans{k_\ell|0, j}{}{}(\cdot | \denoiser{0|j}{}{x_j}, x_j)$}. These surrogates can be computed in closed form for linear inverse problems. For non-linear problems, the authors employ a DPS-like approximation \cite{chung2023diffusion}, i.e., \(\ihpot{j}{x_j}{\ell} \approx \hpot{k_\ell}{\int x_{k_\ell}\, \hat{p} _{k_{\ell}|j}(\rmd x_{k_\ell} \mid x_j)}\).

Then, given these surrogate functions, step (i) is performed using the Tamed ULA algorithm \cite{brosse2019tamed} to mitigate stability issues. Regarding step (ii), for each surrogate transition, samples are generated by first fitting a Gaussian variational approximation and then drawing samples from it.

\subsubsection{Midpoint Guidance Posterior Sampling}
\revised{More recently, \cite{moufad2025variational} proposed the \emph{Midpoint Guidance Posterior Sampling} (MGPS) algorithm 
which relies on sampling approximately from the posterior transition $\post{k|k+1}{x_{k+1}}{}$ \eqref{eq:posterior-backward} at each step of the backward denoising process. It builds on the following decomposition, which holds for all $\ell \in \intset{0}{k-1}$:
\begin{align}
\nonumber
    \post{k|k+1}{x_{k+1}}{\rmd x_k} &\propto \pot{k}{x_k} \pdata{k}{}{\rmd x_k} \fw{k+1|k}{x_k}{x_{k+1}} \\
\nonumber
    &= \int \pot{\ell}{x_\ell} \pdata{\ell|k}{x_k}{\rmd x_\ell} \pdata{k}{}{x_k} \fw{k+1|k}{x_k}{x_{k+1}} \\
\nonumber
    &= \int \pot{\ell}{x_\ell} \pdata{\ell}{}{\rmd x_\ell} \fw{k|\ell}{x_\ell}{x_k} \fw{k+1|k}{x_{k+1}}{x_k} \\
    \label{eq:mgps-decomp}
    &= \int \fw{k|\ell, k+1}{x_{\ell}, x_{k+1}}{x_k} \post{\ell|k+1}{x_{k+1}}{\rmd x_{\ell}} \ \eqsp,
\end{align}
where $\post{\ell|k+1}{x_{k+1}}{\rmd x_{\ell}} \propto \pot{\ell}{x_{\ell}} \pdata{\ell|k+1}{x_{k+1}}{\rmd x_{\ell}}$ and 
\[
\fw{k|\ell, k+1}{x_{\ell}, x_{k+1}}{\rmd x_k} \propto \fw{k|\ell}{x_\ell}{\rmd x_k} \fw{k+1|k}{x_{k}}{x_{k+1}} = \normpdf(x_k; c_{k,\ell} x_\ell+ c_{k,k+1} x_{k+1}; v_{k,\ell})\eqsp, 
\]
where $c_{k,\ell}$, $c_{k,k+1}$, and $v_{k,\ell}$ are constants which can be computed explicitely.
} 
\revised{One step of {\sc{MGPS}} proceeds by first sampling from an approximation of the posterior transition $\post{\ell|k+1}{x_{k+1}}{}$ and then sampling from the bridge transition $\fw{k|\ell, k+1}{x_{\ell}, x_{k+1}}{\cdot}$  to return back to step $k$. The approximation of the posterior transition used in the {\sc{MGPS}} algorithm is 
\begin{equation} 
    \label{eq:mgps-approx}
    \post{\ell|k+1}{x_{k+1}}{x_{\ell}} \approx \frac{\hpot{\ell}{x_{\ell}} \ppdata{\ell|k+1}{x_{k+1}}{x_{\ell}}}{\int \hpot{\ell}{x^\prime _{\ell}} \ppdata{\ell|k+1}{x_{k+1}}{\rmd x^\prime _{\ell}}} \eqsp,
\end{equation}
where $\hpot{\ell}{} = g(\obs|\denoiser{0|\ell}{}{x_{\ell}})$ is the DPS approximation \cite{chung2023diffusion} and $\ppdata{\ell|k+1}{x_{k+1}}{\cdot}$ is a Gaussian approximation of the transition kernel $\pdata{\ell|k+1}{x_{k+1}}{\cdot}$.}

\revised{The time index \(\ell\) typically depends on \(k\), and the choice of \(\ell_k\) governs the trade-off between accurately approximating the likelihood and accurately approximating the Gaussian transition. Specifically, as \(\ell_k\) approaches \(0\), the potential \(\hpot{\ell_k}{}\) becomes a more precise approximation of \(\pot{\ell_k}{}\) defined in \eqref{eq:posterior-forward}, but the Gaussian approximation of the transition deteriorates. For instance, \cite{moufad2025variational} demonstrates, through a solvable toy model, that the choice \(\ell_k \approx \lfloor k/2 \rfloor\) for all \(k\) provides the most accurate approximation of \(\post{}{}{}{}\), whereas the choice \(\ell_k \approx 0\) leads to the least accurate results. The complete MGPS procedure simulates a Markov chain $(\hat{X}_{k})_{k \in \intset{n}{0}}$ where the initial sample is drawn from $\gauss(\zero_\dimx, \Id_\dimx)$ and given $\hat{X}_{k+1}$, the next sample $\hat{X} _k$, which can be considered as an approximate sample from $\post{k| k+1}{x_{k+1}}{}$, is drawn from $\fw{k|\ell_k, k+1}{\tilde{X}_{\ell_k}, \hat{X}_{k+1}}{\cdot}$ with  $\tilde{X}_{\ell_k}$ sampled from $\gauss(\hat{\mu}_{\ell_k}, \hat\Sigma_{\ell_k})$ where $(\hat{\mu}_{\ell_k}, \hat\Sigma_{\ell_k})$ are obtained by performing stochastic gradient descent on 
$$
(\mu, \Sigma) \mapsto \mathsf{KL}\left( \normpdf(\mu, \Sigma) \left\|  \frac{\hpot{\ell_k}{\cdot} \ppdata{\ell_k|k+1}{x_{k+1}}{\cdot}}{\int \hpot{\ell_k}{x^\prime _{\ell_k}} \ppdata{\ell_k|k+1}{x_{k+1}}{\rmd x^\prime _{\ell_k}}}\right. \right) \eqsp.
$$ 
} 
\section{The inpainting problem}
\label{sec:inpainting}
\label{sec:replacement-method}
In this section, we focus specifically on the \emph{inpainting problem}, where $\obs$ is a realization of $Y = A X$, with $A \in \rset^{\dimobs \times \dimx}$, $\dimobs < \dimx$, and $X \sim \prior$. The first algorithm for posterior sampling with a DDM prior was introduced in \cite[Algorithm 2]{song2019generative} and \cite[Appendix I.2]{song2021score}, and following \cite{ho2022video}, we refer to it as the replacement method. For simplicity, we assume in this section that $A = [\Id_{\dimobs \times \dimobs}, \mathbf{O}_{\dimobs \times (\dimx-\dimobs)}]$, meaning we observe the first $\dimobs$ coordinates of $X \sim \prior$. Note that noiseless linear inverse problems can be formulated as an inpainting problem using an SVD.
In this setting, we decompose $x = [\overline{x}, \underline{x}]$, where $\overline{x} \in \rset^\dimobs$ and $\underline{x} \in \rset^{\dimbot}$, with $\dimbot \eqdef \dimx - \dimobs$. Thus, for the remainder of this section, we assume that
\begin{equation}
    \label{eq:inpainting-posterior}
    \post{}{}{\rmd x} = \delta_\obs(\rmd \topx) \prior(\rmd \botx | \topx) 
\end{equation}
where  $\prior(\rmd \botx | \topx)$ is a Markov kernel that satisfies $\prior(\rmd x) = \prior(\rmd \botx | \topx) \bprior(\rmd \topx)$.
In other words, we observe the first $\dimobs$ coordinates and want to infer the $\dimbot$ remaining ones by leveraging the information from the prior $\prior$. In what follows we denote by $\tfw{k|0}{\topx_0}{\cdot}$ and $\bfw{k|0}{\botx_0}{\cdot}$ the marginals of  $\fwtrans{k|0}{x_0}{\cdot}$ \wrt\ to the first $\dimobs$ and last $\dimbot$ coordinates respectively; e.g. we have that $\tfw{k|0}{\topx_0}{\topx_k} \eqdef \normpdf(\topx_k; \sqrt{\acp{k}}\, \topx_0, \var_{k|0} \Id_\dimobs)$ and $\fwtrans{k|0}{x_0}{x_k} = \tfw{k|0}{\topx_0}{\topx_k} \bfw{k|0}{\botx_0}{\botx_k}$. In such a setting, the target distribution (see \eqref{eq:target-posterior} and \eqref{eq:posterior-forward}) writes
\begin{equation}
\label{eq:target-posterior-inpainting}
\post{k}{}{\rmd x_k}= \post{k}{}{\rmd (\topx_k, \botx_k)} = \tfw{k|0}{\obs}{\rmd \topx_k} \int  \prior(\rmd \botx_0 | \obs) \bfw{k|0}{\botx_0}{\rmd \botx_k} = \frac{\fwmarg{k}{\rmd x_k} \tfw{0|k}{x_k}{\obs}}{\fwmarg{k}{}(\tfw{0|k}{\cdot}{\obs})} .
\end{equation}
where we have set, using the definition \eqref{eq:posterior-forward} of the backward kernel 
\begin{equation}
\label{eq:definition-tfw}
\tfw{0|k}{x_k}{\obs} = \int \pdata{0|k}{x_k}{[\obs,\botx_0]} \rmd \botx_0 .
\end{equation}
\subsection{The Replacement method}
\label{subsec:replacement-method}
The \emph{replacement method} defines a backward Markov chain \((X_k)_{k=n}^0\) as follows. First, sample \(X_n \sim \pdata{n}{}{}\). Then, recursively, given \(X_{k+1}\), sample \(\tilde{X}_k \sim \pdata{k|k+1}{X_{k+1}}{\cdot}\) from the unconditional transition. A pseudo-observation \(Y_k\) is then constructed as \(Y_k \sim \tfw{k|0}{\obs}{\cdot}\). Next, one sets $X_k = [Y_k, \underline{\tilde{X}}_k]$ and proceeds to the next step. Put simply, this method iteratively performs the denoising diffusion process, but at each step $k$ the first $\dimobs$ coordinates are replaced with an exact sample $Y_k$ from the marginal w.r.t. the first $\dimobs$ coordinates of $\post{k}{}{}$ (see \eqref{eq:target-posterior-inpainting}). Although the replacement method approach yields satisfactory results on image inpainting problems - see, e.g., \cite[Corollary D.2]{trippe2023diffusion}, it does not sample, even approximately, the target distribution.

 An interpretation of the replacement method can be drawn by considering the following sequence of distributions \((\hpost{k}{}{})_{k=0}^n\), where \(\hpost{n}{}{} \eqdef \pdata{n}{}{}\), and for each \(k \in \intset{1}{n-1}\), \(\hpost{k}{}{}\) is defined as
\begin{equation}
\label{eq:replacement-sequence}
\hpost{k}{}{x_k} = \tfw{k|0}{\obs}{\topx_k} \bpdata{k}{\topx_k}{\botx_k}, \quad \text{where} \quad \bpdata{k}{\topx_k}{\botx_k} \eqdef \frac{\pdata{k}{}{x_k}}{\tpdata{k}{}{}(\topx_k)}= \frac{\pdata{k}{}{[\topx_k, \botx_k]}}{\tpdata{k}{}{\topx_k}} 
\end{equation}
with $\tpdata{k}{}{}(\topx_k)= \int \pdata{k}{}{}([\topx_k, \botx^\prime _k]) \rmd \botx^\prime _k$ and \(\tfw{k|0}{\obs}{\topx_k} \eqdef \normpdf(\topx_k; \sqrt{\acp{k}} \obs, \var_{k|0} \Id_\dimobs)\).
Compared to \eqref{eq:target-posterior-inpainting}, the distribution of the first $\dimobs$-components (\( \tfw{k|0}{\obs}{\rmd \topx_k}\))  is exact but the distribution of the last $(d-\dimobs)$-components (\(\bpdata{k}{\topx_k}{\botx_k}\)) is inexact.  
We now show that the transition of the replacement method corresponds to one iteration of a Gibbs sampler targeting \(\hpost{k}{}{}\).
First, sampling from $\hpost{k}{}{}$ is equivalent to sampling $\topX_k \sim \tfw{k|0}{\obs}{\cdot}$ and then $\botX_k \sim \bpdata{k}{\topX_k}{\cdot}$. Since the sampling of $\topX_k$ is exact, we only need to focus on sampling $\botX_k$. Next, decomposing the transition $\pdata{k|k+1}{x_{k+1}}{\cdot}$ \wrt\ the top and bottom coordinates yields 
$$ 
    \pdata{k|k+1}{x_{k+1}}{x_k} = \bpdata{k|k+1}{x_{k+1}, \topx_{k}}{\botx_k} \tpdata{k|k+1}{x_{k+1}}{\topx_k}
$$
where 
\begin{align*} 
    \tpdata{k|k+1}{x_{k+1}}{\topx_k} & \eqdef \int \pdata{k|k+1}{x_{k+1}}{[\topx_k, \botx_k]} \, \rmd \botx_k \eqsp, \\
    \bpdata{k|k+1}{x_{k+1}, \topx_k}{\botx_k} & \eqdef \pdata{k|k+1}{x_{k+1}}{[\topx_k, \botx_k]} \big/ \tpdata{k|k+1}{x_{k+1}}{\topx_k} \eqsp.
\end{align*}
It then follows that 
\begin{align*} 
    \pdata{k}{\topx_k}{\botx_k} & = \frac{\int \bpdata{k|k+1}{x_{k+1}, \topx_k}{\botx_k} \tpdata{k|k+1}{x_{k+1}}{\topx_k} \pdata{k+1}{}{\rmd x_{k+1}}}{\tpdata{k}{}{\topx_k}} \\
    & = \int \bpdata{k|k+1}{x_{k+1}, \topx_k}{\botx_k} \tpdata{k+1}{\topx_k}{\rmd x_{k+1}}
\end{align*}
where $\tpdata{k+1}{\topx_k}{\rmd x_{k+1}} \eqdef \pdata{k+1}{}{\rmd x_{k+1}} \tpdata{k|k+1}{x_{k+1}}{\topx_k} / \tpdata{k}{}{\topx_k}$. 
Hence, for all $\topx_k \in \rset^{\dimbot}$, $\bpdata{k}{\topx_k}{\cdot}$, defined in \eqref{eq:replacement-sequence}, is the marginal \wrt\ the first variable of the joint distribution with density 
\begin{equation}
    \label{eq:repl-bw}
    \bpdata{k, k+1}{\topx_k}{\botx_k, x_{k+1}} = \bpdata{k|k+1}{x_{k+1}}{\botx_k} \tpdata{k+1|k}{\topx_k}{x_{k+1}}
\end{equation}
As a result, approximate samples can be obtained by using a Gibbs sampler \cite{gelfand1990sampling} targeting $\smash{\bpdata{k, k+1}{\topx_k}{\rmd (\botx_k, x_{k+1})}}$, and retaining only the $\botx_k$-coordinate from the generated samples. In this context, the Gibbs sampler constructs a Markov chain $(\botX^r_k, X^r_{k+1})_{r=1}^{R}$ by alternately sampling from the full conditionals of \eqref{eq:repl-bw}. The density of the conditional distribution of $\botX_k$ given $X_{k+1} = x_{k+1}$ is $\bpdata{k|k+1}{x_{k+1}, \topx_k}{\cdot}$ and that of $X_{k+1}$ given $(\topX_k, \botX_k) = (\topx_k, \botx_k)$ is simply the forward transition $\fw{k+1|k}{[\topx_k, \botx_k]}{\cdot}$. As for the first full conditional, it can be sampled approximately by noting that the DDPM Gaussian transition $\ppdata{k|k+1}{x_{k+1}}{\cdot}$ \eqref{eq:ddpm-kernel} has diagonal covariance and as such, we have that 
\begin{equation}
    \label{eq:ddpm-decomp}
    \pdata{k|k+1}{x_{k+1}}{x_k} \approx \tpdata{k|k+1}{x_{k+1}}{\topx_k} \bpdata{k|k+1}{x_{k+1}}{\botx_k} 
\end{equation}
where $\tpdata{k|k+1}{x_{k+1}}{\cdot}$ and $\bpdata{k|k+1}{x_{k+1}}{\cdot}$ are the marginals of $\ppdata{k|k+1}{x_{k+1}}{\cdot}$ \wrt\ to the top and bottom coordinates. Thus, under the approximate model, $\bpdata{k|k+1}{x_{k+1}, \topx_k}{\cdot} \approx \bpdata{k|k+1}{x_{k+1}}{}$, \emph{i.e.} there is no dependence on $\topx_k$. 
 To summarize, the Gibbs sampler we have just described operates as follows; given an initial sample $X^0 _{k+1} = X_{k+1}$ where $X_{k+1}$ is an approximate sample from the previous distribution $\post{k+1}{}{}$ in the sequence and $\topX_k = \sqrt{\acp{k}} \obs + \sqrt{1 - \acp{k}} Z_k$ where $Z_k \sim \gauss(\zero_\dimobs, \Id_\dimobs)$, repeat the updates for $r \in \intset{1}{R}$,
\begin{equation}
    \label{eq:replacement:gibbs}
        \begin{cases}
            \botX^{r} _k \sim \bpdata{k|k+1}{X^{r-1} _{k+1}}{\cdot} \eqsp,\\
            X^{r} _{k+1} \sim \fwtrans{k+1|k}{[\topX_k, \botX^{r} _k]}{\cdot} \eqsp. 
        \end{cases} 
    \end{equation}
Setting \(R = 1\) exactly recovers the replacement method described above.

The procedure we outlined bears similarities to the \emph{RePaint} algorithm \cite{lugmayr2022repaint}. In our updates \eqref{eq:replacement:gibbs}, the top coordinate is first sampled from \(\tfw{k|0}{\obs}{\cdot}\) and remains fixed throughout the subsequent iterations. In contrast, RePaint refreshes this coordinate at each iteration, \emph{i.e.}, its updates are instead:
\begin{equation}
    \label{eq:replacement:repaint}
        \begin{cases}
            \topX^r _k \sim \tfw{k|0}{\obs}{\cdot} \eqsp, \\
            \botX^{r} _k \sim \bpdata{k|k+1}{X^{r-1} _{k+1}}{\cdot} \eqsp,\\
            X^{r} _{k+1} \sim \fwtrans{k+1|k}{[\topX^r _k, \botX^{r} _k]}{\cdot} \eqsp,
        \end{cases} 
\end{equation}
see \cite[Algorithm 1]{lugmayr2022repaint} for details. Interestingly, the authors observed that increasing \(R\) enhances the algorithm's performance, as demonstrated in \cite[Figure 3 and Table 3]{lugmayr2022repaint}. This finding is consistent with the theoretical principles of Gibbs sampling, despite the fact that RePaint itself is not a Gibbs sampler. 
\subsection{SMCDiff: Sequential Monte Carlo Diffusion algorithm}
The \emph{SMCDiff}, which was proposed in \cite{trippe2023diffusion}, also works with the sequence \eqref{eq:replacement-sequence}, but uses an SMC approach. However, there is a significant difference to the approach considered in \Cref{sec:sequence-distribution}-\ref{subsec:smc}, as the particle approximation is not applied to $\hpost{k}{}{}$ in \eqref{eq:replacement-sequence}. Indeed, the decomposition \eqref{eq:replacement-sequence} shows that the ``top'' coordinates $(\topX_k)_{k=1}^n$ can be simulated precisely by simply applying the forward kernel: more precisely, we initialize with $\topX_0 = \obs$, then recursively sample from the marginal forward kernel, $\topX_{k+1} \sim \tfw{k+1|k}{\topX_k}{\cdot}$. 
This method is similar to the replacement method, as it samples pseudo-observations for the "top" components. The idea is to show now that the remaining component, the sequence of distributions
$\bpdata{k}{\topx_k^*}{\botx_k}$, is defined by a recursion of the form \eqref{eq:FK-semigroup}, thereby defining a Feynman-Kac semigroup, which will allow this sequence of distributions to be approximated by a particle system. We use the notation $(\topx_k^*)_{k=1}^n$ to stress that these components are frozen.  
It holds for all $(\topx^*_k, \topx^*_{k+1}) \in \rset^{\dimobs}$ that 
$$
\tpdata{k}{\topx^*_k}{\botx_k} = \int \frac{\pdata{k|k+1}{[\topx^*_{k+1},\botx_{k+1}]}{[\topx^*_k,\botx_k]} \pdata{k+1}{}{}([\topx^*_{k+1},\botx_{k+1}])}{\int \pdata{k|k+1}{[\topx^*_{k+1}, \botx^\prime_{k+1}]}{[\topx^*_k, \botx^\prime _k]} \pdata{k+1}{}{}([\topx^*_{k+1}, \botx^\prime _{k+1}]) \rmd \botx^\prime _{k:k+1}} \rmd \botx _{k+1} \eqsp.
$$
Finally, with \eqref{eq:ddpm-decomp} and dividing both the numerator and denominator by $\tpdata{k+1}{}{}(\topx^*_{k+1})$, we obtain the recursion
$$
    \bpdata{k}{\topx^*_k}{\botx_k} = \frac{1}{\mathrm{Z}_{k+1}^*}\int \bpdata{k|k+1}{[\topx^*_{k+1},\botx_{k+1}], \topx^* _k}{\botx_k} \tpdata{k|k+1}{[\topx^*_{k+1},\botx_{k+1}]}{\topx^*_k} \bpdata{k+1}{\topx^*_{k+1}}{\rmd \botx_{k+1}}
$$
where $\mathrm{Z}_{k+1}^* \eqdef \int \tpdata{k|k+1}{[\topx^*_{k+1}, \botx^\prime _{k+1}]}{\topx^*_k} \, \bpdata{k+1}{\topx^*_{k+1}}{\rmd \botx_{k+1}}$. This recursion is of the form \eqref{eq:FK-semigroup} and an SMC algorithm can be applied to it once we plugged the parametric model. Given a particle approximation of $\bpdata{k+1}{\topx^*_{k+1}}{\cdot}$, \cite{trippe2023diffusion} obtain the next particle approximation of $\bpdata{k}{\topx^*_k}{\cdot}$ using a {\sc weight-resample-shake} scheme with adjustment multiplier function $\adjopt_{k+1}(\botx_{k+1}) = \tpdata{k|k+1}{[\topx_{k+1}^*,\botx_{k+1}]}{\topx^*_k}$ and proposal $\ropt_{k|k+1}(\cdot | \botx_{k+1}) = \bpdata{k|k+1}{[\topx^*_{k+1},\botx_{k+1}]}{\cdot}$. It is interesting to note that even when the number of particles tends to infinity, this algorithm does not yield a weighted sample that consistently estimates \( \hpost{k}{}{} \), because the particle system remains conditional on a single trajectory of the pseudo-observations. 

\subsection{MCGDiff for inpainting}
\label{subsec:mcgdiff}
The MCGDiff algorithm for the inpainting problem was proposed in \cite{cardoso2024monte}. It is also based on \eqref{eq:target-posterior-inpainting} and is conceptually similar to SMCDiff, but uses a different sequence of distributions:
\begin{equation}
    \label{eq:mcgdiff-sequence}
    \hpost{k}{}{x_k} = \tfw{k|0}{\obs}{\topx_k} \pdata{k}{}{x_k} \big/ \int \tfw{k|0}{\obs}{\topx^\prime _k} \pdata{k}{}{\rmd \topx^\prime _k} 
\end{equation}
A comparison of this distribution with \eqref{eq:target-posterior-inpainting} shows that the unconditional marginal distribution \( \fwmarg{k}{} \) has been replaced by its diffusion approximation \( \ppdata{k}{}{} \), which is perfectly in line with the spirit of DDM, and the likelihood term \( \tfw{0|k}{x_k}{\obs} \) by \( \tfw{k|0}{\obs}{\topx_k} \). Compared to 
\eqref{eq:replacement-sequence}, 
\eqref{eq:mcgdiff-sequence}  uses  the posterior information $\ppdata{k}{}{}$ on both coordinates instead of only the bottom ones through $\bpdata{k}{\topx_k}{\cdot}$. It is easily seen that \eqref{eq:mcgdiff-sequence} satisfies the recursion \eqref{eq:FK-semigroup} with 
\begin{equation*}
    w_k(x_k,x_{k+1})  = \frac{\tfw{k|0}{\obs}{\topx_{k}}}{ \tfw{k+1|0}{\obs}{\topx_{k+1}}} \eqsp, \quad 
    m_k(x_k | x_{k+1})  = \pdata{k|k+1}{x_{k+1}}{x_k} \eqsp,
\end{equation*}
and hence, \textsc{resample--shake--weight} can be used to  obtain an approximation of each distribution $\post{k}{}{}$. The optimal proposal kernel $\smash{\ropt_{k|k+1}(x_k | x_{k+1}) \propto \tfw{k|0}{\obs}{\topx_k} \pdata{k|k+1}{x_{k+1}}{x_k}}$ and adjustment multipliers $\adjopt_{k+1}(x_{k+1}) = \int \tfw{k|0}{\obs}{\topx_k} \pdata{k|k+1}{x_{k+1}}{\rmd x_k} / \tfw{k+1|0}{\obs}{\topx_{k+1}}$ are both available in closed form once the parametric model is used. \cite{cardoso2024monte} proposed in such case to implement the {\sc weight-resample-shake} scheme described in \Cref{sec:sequence-distribution}\ref{subsec:smc}. 
Compared to \cite{trippe2023diffusion}, the resulting approximation consists of $N$ particles for both the top and bottom coordinates. Finally, while in the previous methods the top coordinates are sampled exactly according $\tfw{k|0}{\obs}{\cdot}$, here they are, conditionally on $X_{k+1} = x_{k+1}$, sampled from the conditional distribution  $\tropt_{k|k+1}(\rmd \topx_k | x_{k+1}) \propto \tfw{k|0}{\obs}{\rmd \topx_k} \tpdata{k|k+1}{x_{k+1}}{\topx_k}$.
Denote $\tbwmean{k|k+1}(x_{k+1})= \int \topx_k \,\tpdata{k|k+1}{x_{k+1}}{\rmd \topx_k}$. A sample $\topX_{k}$ from $\tropt_{k|k+1}(\cdot | x_{k+1})$ writes 
$$ 
    \topX_{k} = \frac{\var^\param _{k|k+1}}{\var _{k|0} + \var^\param _{k| k+1}} \sqrt{\acp{k}}  \obs + \frac{\var_{k|0}}{\var _{k|0} + \var^\param _{k|k+1}}\tbwmean{k|k+1}(x_{k+1}) + \var^\obs _{k|k+1}\overline{Z} _k 
$$ 
where $\smash{\var^\obs _{k|k+1} \eqdef [\var^\param _{k| k+1} \var _{k|0} / (\var^\param _{k| k+1} + \var _{k|0})]^{1/2}}$.  As we compute a convex combination of $\sqrt{\acp{k}} \obs$ and $\tbwmean{k|k+1}(x_{k+1})$ before injecting noise, this update rather corresponds to a \emph{soft} replacement method. 

\cite{cardoso2024monte} also considers linear inverse problems with Gaussian noise, i.e. $\pot{}{x} = \normpdf(\obs; A x, \sigma^2_\obs \Id)$ with $\sigma^2_\obs > 0$ and shows how the noiseless inpainting methods can be extended to noisy setting. Let us again consider the inpainting problem, but with noise, i.e. $\pot{}{x} = \normpdf(\obs; \topx, \sigma_\obs^2 \Id)$. Further assume that there exists $\tau \in \intset{0}{n}$ so that $\sigma^2 _\obs = (1 - \acp{\tau}) / \acp{\tau}$. Then, it holds that $\pot{}{x} = \tfw{\tau|0}{\topx}{\sqrt{\acp{\tau}} \obs}$, which in turn allows us to write 
\begin{align*} 
    \post{}{}{\rmd x} \propto \pot{}{x} \prior(\rmd x) & \propto \int \prior(\rmd x) \tfw{\tau|0}{\topx}{\sqrt{\acp{\tau}} \obs} \bfw{\tau|0}{\botx}{\botx_\tau} \, \rmd \botx_\tau \\
    & \propto \int \prior(\rmd x) \fwtrans{\tau|0}{x}{x_\tau} \delta_{\sqrt{\acp{\tau}} \obs}(\rmd \topx_\tau) \, \rmd \botx_\tau \eqsp.
\end{align*}
 Using then the identity $\prior(\rmd x) \fwtrans{\tau|0}{x}{x_\tau} = \fwmarg{\tau}{x_\tau} \pdata{0|\tau}{x_\tau}{\rmd x}$, we see that 
$$ 
    \post{}{}{\rmd x} = \int \pdata{0|\tau}{x_\tau}{\rmd x} \tilde\pi^\obs _\tau(\rmd x_\tau) \eqsp, \quad \mathrm{where} \quad \tilde\pi^\obs _\tau(\rmd x_\tau) \propto \delta_{\sqrt{\acp{\tau}} \obs}(\rmd \topx_\tau) \fwmarg{\tau}{x_\tau} \rmd \botx_\tau \eqsp.
$$ 
which in turn shows that to sample from $\post{}{}{}$, one first needs to sample $X_\tau \sim \tilde\pi_\tau$ and then $X_0 \sim \pdata{0|\tau}{X_\tau}{\cdot}$. The former step corresponds to an inpainting problem with the rescaled observation $\sqrt{\acp{\tau}} \obs$, prior $\fwmarg{\tau}{}$, which is approximated by $\ppdata{\tau}{}{}$, and can be handled with a straightforward adaptation of MCGDiff. See \cite[Appendix A]{cardoso2024monte} for more details. 

We end this section by noting that \emph{approximate} Gibbs samplers, as described in the previous sections, could be used to sample from the sequence of  distributions used in \cite{cardoso2024monte} and \cite{dou2024diffusion} by alternately sampling the conditionals $m_{k|k+1}$ and $\fwtrans{k+1|k}{}{}$.

\section{Further comments and perspectives}

We conclude with important considerations and questions arising from the preceding discussion.
\paragraph{Choosing the distribution sequence.}
One of the main challenges is identifying a method for comparing different sequences of distributions. Although it is clear that consecutive distributions must remain close for efficient sampling, quantifying this proximity precisely is still an open problem. Important considerations include selecting the most suitable divergence measure or distance metric and evaluating whether the sequence \(\post{k}{}{}\) indeed represents the optimal target distribution. If so, another critical question emerges: how should neural networks be trained to accurately approximate the associated potentials \(\pot{k}{}\)? Recent studies \cite{phillips2024particle,denker2024deft,li2024derivative} have explored various approaches for learning these potentials, but consensus on their improvements over established methods has yet to be reached. A promising alternative is offered by \cite{bruna2024posterior}, who propose designing potentials specifically for linear inverse problems by enforcing a suitable Fokker–Planck equation, resulting in a theoretically grounded framework.

\paragraph{Choosing the best sampling scheme.}  A second open question relates to selecting an appropriate sampling design, which typically comprises three main components. Existing methods in the literature predominantly rely on Sequential Monte Carlo (SMC), Gibbs sampling, Langevin dynamics, or non-amortized variational inference (VI), each having distinct strengths and weaknesses. Determining which of these methods provides the best trade-off remains unclear. Additionally, while non-amortized VI has shown promise—particularly for parallel sample generation—its efficiency and scalability compared to more established methods, such as SMC or Langevin dynamics, have yet to be fully understood.

\paragraph{Efficiency of DPS-type potentials.}
Empirical evidence, notably presented in the comprehensive benchmark by \cite{janati2025mixture}, indicates that DPS-type potentials outperform alternative approaches in terms of Frechet Inception Distance and Learned Perceptual Image Patch Similarity metrics. However, these improvements involve important trade-offs: DPS-based methods are computationally slower and significantly more memory-intensive, limiting their scalability and practical usability in resource-constrained environments. A critical open question is whether vector Jacobian product-free methods could match DPS performance while reducing computational costs. 

\bibliographystyle{plain}
\bibliography{bibliography}

\end{document}